\documentclass[10pt,twocolumn,letterpaper]{article}

\usepackage{iccv}              

\newcommand{\mr}[2]{\multirow{#1}{*}{#2}}
\newcommand{\mc}[2]{\multicolumn{#1}{c}{#2}}
\newcommand{\fst}[1]{\textcolor{blue}{#1}}
\newcommand{\snd}[1]{\textcolor{magenta}{#1}}
\newcommand{\mbf}[1]{\mathbf{#1}}

\newcommand{\ph}[1]{\vspace{1mm}\noindent{\textbf{#1}}}
\newcommand{\dataset}[1]{\textsc{#1}}
\DeclareMathOperator{\diag}{diag}
\DeclareMathOperator{\sm}{softmax}

\newcommand{\ours}{\textit{Concertormer}}
\newcommand{\ourslite}{\textit{Concertormer-lite}}
\newsavebox{\measurebox}

\definecolor{iccvblue}{rgb}{0.21,0.49,0.74}
\usepackage[pagebackref,breaklinks,colorlinks,allcolors=iccvblue]{hyperref}

\usepackage{amsmath}
\usepackage{bbding}
\usepackage{multirow}
\usepackage{paralist}
\usepackage{tablefootnote}
\usepackage[table]{xcolor} 
\def\paperID{11983} 
\def\confName{ICCV}
\def\confYear{2025}

\title{Efficient Concertormer for Image Deblurring and Beyond}

\author{Pin-Hung Kuo\textsuperscript{1}
\and
Jinshan Pan\textsuperscript{2}
\and 
Shao-Yi Chien\textsuperscript{1}
\and
Ming-Hsuan Yang\textsuperscript{3}\\ \vspace{-13pt} \and
\textsuperscript{1}National Taiwan University \and
\textsuperscript{2}Nanjing University of Science and Technology \and
\textsuperscript{3}University of California, Merced \and
}
\begin{document}
\maketitle
\begin{abstract}
The Transformer architecture has excelled in NLP and vision tasks, but its self-attention complexity grows quadratically with image size, making high-resolution tasks computationally expensive. We introduce {\ours}, featuring Concerto Self-Attention (CSA) for image deblurring. CSA splits self-attention into global and local components while retaining partial information in additional dimensions, achieving linear complexity. A Cross-Dimensional Communication module enhances expressiveness by linearly combining attention maps. Additionally, our gated-dconv MLP merges the two-staged Transformer design into a single stage. Extensive evaluations show our method performs favorably against state-of-the-art works in deblurring, deraining, and JPEG artifact removal. Code and models are available at \href{https://github.com/setsunil/Concertormer}{https://github.com/setsunil/Concertormer}. 
\end{abstract}    
\vspace{-20pt}
\section{Introduction}
\label{sec:intro}
In recent years, deep learning has driven significant advances in image restoration, with deep convolutional neural networks (CNNs) becoming widely used for image deblurring. The core operation in CNNs, i.e., convolution, is effective for extracting local features but limited in capturing non-local features, which are essential for deblurring.

The self-attention (SA) mechanism, on the other hand, excels at capturing non-local information and has shown promising results in image deblurring \cite{chen2021pre}. 
However, the high computational costs make it challenging to apply, particularly to high-resolution images.

\begin{figure}[!t]
  \centering
\includegraphics[width=1.\linewidth]{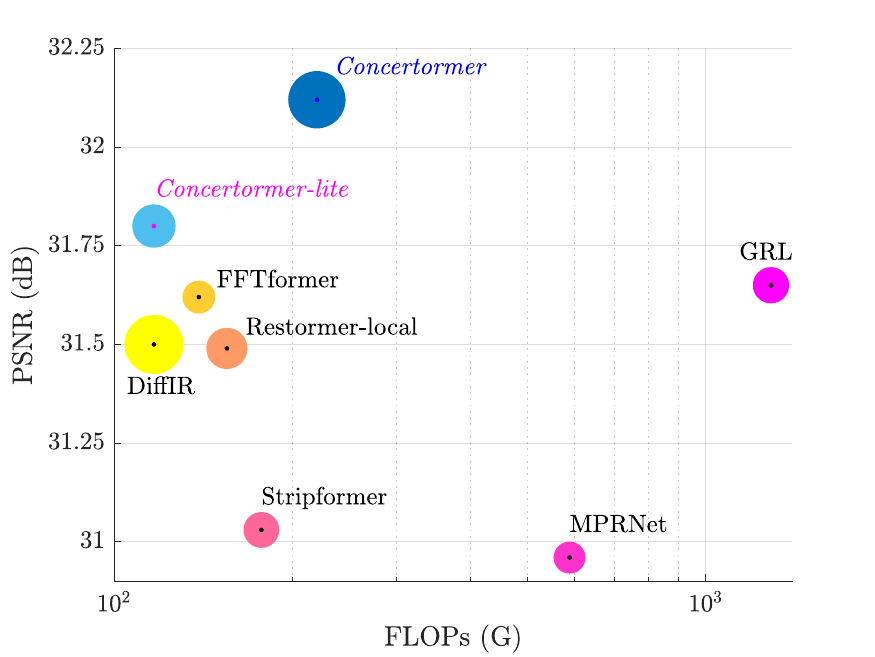}
\vspace{-6mm}
 \caption{PSNR vs FLOPs on the \dataset{HIDE} \cite{shen2019human} dataset. Our models are marked in \textit{\fst{blue}} and \textit{\snd{magenta}} text. The area of each circle is proportional to the number of parameters.}
 \label{fig:flops}
\end{figure}

\begin{figure*}[tp]
\vspace{-5pt}
    \centering
    \includegraphics[width=0.96\linewidth]{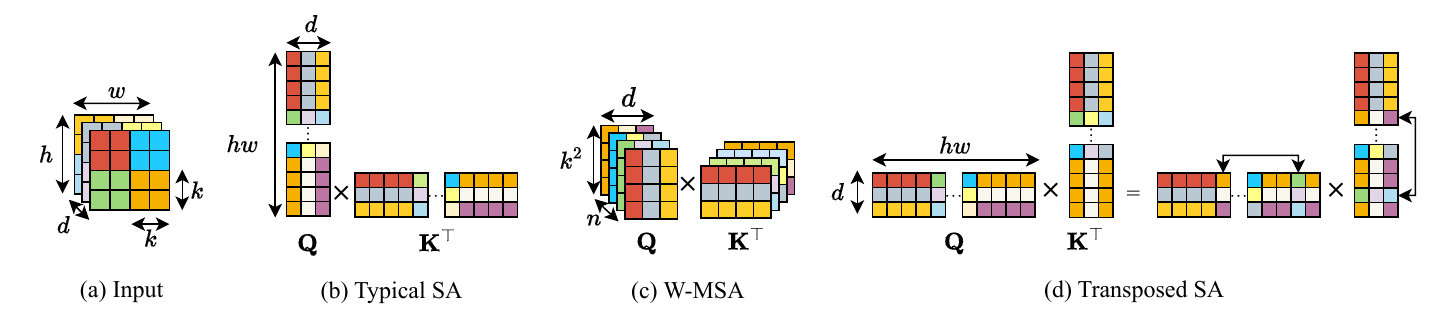}
    \vspace{-3mm}
    \caption{Self-attention methods. 
    (a) Input tensor. (b) Typical self-attention. (c) Window multi-head self-attention \cite{liu2021swin, liu2022swin}, where $n=hw/k^2$ is the number of blocks. (d) Transposed self-attention \cite{zamir2022restormer} and a random column permutation. Column permutations do not affect the resulting attention map.}
    \label{fig:attn}
    \vspace{-4mm}
\end{figure*}

Window-based multi-head self-attention (W-MSA) \cite{liang2021swinir, wang2022uformer} mitigates computational costs by limiting the calculation area. However, this approach suffers from limited receptive fields, making it less effective for capturing global relationships.
As illustrated in Fig.~\ref{fig:attn}(c), W-MSA processes each block in the input feature map independently. 
Although the shifted window technique communicates attention between neighboring blocks, it requires a sufficient number of blocks to model global relationships effectively.

Transposed SA is an alternative approach, as used in Restormer \cite{zamir2022restormer}, which calculates self-attention along the channel dimension instead of the spatial dimension.
However, this approach underestimates the importance of local connectivity. As shown in Fig.~\ref{fig:attn}(d), transposed self-attention loses spatial connectivity during matrix multiplication: random column permutations in $\mathbf{Q}$ and $\mathbf{K}$ do not affect the result, meaning that transposed SA scarcely retains spatial connectivity information.

To address the aforementioned problems, we propose an efficient Transformer of Concerto Self-Attention, i.e., {\ours}, specifically for image deblurring. This model simultaneously captures both local and global connectivity with a lower computational cost. By dividing feature maps into grids and a specific number of heads, Concerto Self-Attention introduces an additional dimension compared to the typical multi-head self-attention mechanism.
To utilize this additional dimension along with the attention heads, we introduce a learnable linear projection mechanism, termed \textit{Cross-Dimensional Communication}, which enables aggregation across independent attention maps.

The CSA is an efficient framework for self-attention calculation, which is adaptable to existing methods, such as transposed SA for image restoration. Notably, the computational complexity of this self-attention mechanism scales linearly with image size (Section~\ref{sec:MA}).

Another key component of Transformer models is the feed-forward network (FFN), which functions as a key-value memory mechanism \cite{geva2020transformer} and is essential for contextualization \cite{kobayashi2023analyzing}. This has led to the widespread use of a two-stage architecture, i.e., self-attention followed by an FFN, in NLP applications.
Although the FFN's role in enhancing feature representation is well-established in NLP, its effectiveness in vision tasks remains uncertain. Moreover, the conventional two-stage Transformer architecture limits design flexibility. To address this, we introduce the gated-dconv MLP (gdMLP), an adaptation of the gated MLP \cite{liu2021pay} with depth-wise convolution and without layer normalization, condensing the two-stage structure into a single stage and reducing model complexity.

The main contributions of this work are as follows:
\begin{compactitem}
\item We propose \textit{Concerto Self-Attention} to effectively model both global and local features across spatial and channel domains with linear complexity.
\item Leveraging the additional dimensions provided by Concerto Self-Attention, we introduce a \textit{Cross-Dimensional Communication} module that linearly combines independent attention maps, enhancing feature exploration for improved image restoration.
\item We develop a streamlined \textit{gdMLP} module, which simplifies the FFN in Transformers, thereby enhancing architectural flexibility and expressive ability.
\end{compactitem}
We evaluate the proposed method on image deblurring datasets, demonstrating its favorable performance against state-of-the-art approaches. Additionally, we show that it is versatile, effectively extending to other image restoration tasks like deraining and deblurring with JPEG artifacts.

\section{Related Work}
\label{sec:rw}
\ph{Image Deblurring.} Traditionally, blind image deblurring has relied on developing image priors. Inspired by dehazing methods \cite{he2010single}, the dark channel prior proved effective for deblurring \cite{pan2016blind}. Since then, various priors have been proposed, including the hyper-Laplacian prior \cite{pan2016robust}, extreme channels prior \cite{yan2017image}, graph-based image prior \cite{bai2018graph}, local maximum gradient prior \cite{chen2019blind}, and sparsity-based methods \cite{chen2020enhanced}.
Recently, CNNs \cite{lai2017deep, zhang2018image, anwar2020densely, zamir2020learning, zhang2020residual} have outperformed conventional priors in image restoration tasks. Multi-scale design became popular in computer vision \cite{he2017multi, hu2017fastmask, park2017unified, lai2017deep, yang2017high, lai2018fast} and was soon adopted for deblurring. Nah et al. \cite{nah2017deep} introduced a multi-scale model with residual blocks \cite{he2016deep}. Following U-Net's success in segmentation \cite{ronneberger2015u}, researchers adopted this architecture for deblurring, utilizing its hierarchical structure to explore cross-scale relationships \cite{tao2018scale, kupyn2019deblurgan, zhang2019deep, abuolaim2020defocus, cho2021rethinking, zamir2021multi, zhang2021plug, chen2022simple}. Tao et al. \cite{tao2018scale} refined this multi-scale design by integrating U-Nets with LSTMs as latent layers, while Zhang et al. \cite{zhang2019deep} proposed a multi-patch network where each U-Net processes differently sized patches, inspiring further work \cite{zamir2021multi}.

Rather than using cascaded U-Nets for different resolutions or patches, Cho et al. \cite{cho2021rethinking} integrate multi-scale design into a single U-Net architecture, and, to the best of our knowledge, are the first to apply a frequency-domain loss function for deblurring. Instead of modifying the structure of U-Net, Chen et al. \cite{chen2022simple} introduce a building unit based on a gated architecture \cite{dauphin2017language} and a squeeze-and-excitation block \cite{hu2018squeeze}, resulting in an efficient and effective U-Net.

\ph{Vision Transformers.} Recognizing the importance of global dependencies in high-level vision tasks, numerous vision Transformers (ViTs) have been developed for applications such as recognition \cite{dosovitskiy2020image, touvron2021training, yuan2021tokens}, video and image classification \cite{wang2018non, liu2021swin, liu2022swin}, object detection \cite{carion2020end, zhu2020deformable, zhang2021learning}, and semantic segmentation \cite{zheng2021rethinking, strudel2021segmenter, xu2022multi}.
In low-level vision tasks, the high-resolution inputs make typical self-attention computationally costly. To address this, efficient variants such as W-MSA \cite{liang2021swinir, wang2022uformer}, transposed self-attention \cite{zamir2022restormer}, stripe and anchored stripe attention \cite{tsai2022stripformer, li2023efficient}, and frequency-based attention \cite{kong2023efficient} have been developed. 
While these approaches reduce complexity from quadratic to linear, they often sacrifice key characteristics of standard self-attention. For example, W-MSA and stripe attention restrict the self-attention region, weakening global contextual representation, and frequency-based attention is mathematically closer to convolution than self-attention.
In contrast to these methods, we introduce Concerto Self-Attention, a mechanism designed to capture both local and global relationships with linear complexity. Our approach achieves the highest PSNR and SSIM scores with relatively low FLOPs among deblurring Transformers.
\section{Revisiting Self-Attention}
\label{sec:revisit}
To better motivate our work, we begin by revisiting typical self-attention, W-MSA, and transposed self-attention. 
For typical self-attention, it first constructs \textit{query} $Q$, \textit{key} $K$ and \textit{value} $V$ from the normalized tensor $\mbf{X} \in \mathbb{R}^{h\times w\times d}$ as: 
\begin{align}
\label{eq:qkv}
    Q =  \mbf{W}_p^q(\mbf{X}),
    K =  \mbf{W}_p^k(\mbf{X}),
    V =  \mbf{W}_p^v(\mbf{X}),
\end{align}
where $Q, K, V \in \mathbb{R}^{hw\times d}$, $\mbf{W}_p^q$, $\mbf{W}_p^k$ and $\mbf{W}_p^v$ are linear projection operations, and $_p$ denotes a pixel-wise operation.
Then the self-attention is
\begin{equation}
\text{SA}(\mbf{X})=\sm(QK^\top/\sqrt{d})V,
    \label{eq:sa}
\end{equation}
where the $\sm(\cdot)$ performs to the last dimension and $d$ is the vector length of inner product.

Although the exhaustive calculation of self-attention yields satisfying results, the complexity is too high to be used for high-resolution images. W-MSA is proposed as an efficient substitute based on an intuitive assumption: Most large attention values are concentrated on the diagonal of an attention map because neighbor pixels are supposed to be highly correlated. W-MSA firstly divides the $Q, K, V$ into $k\times k$ non-overlapping blocks: $Q_i, K_i, V_i \in \mathbb{R}^{k^2\times d}$, where $1\leq i\leq \frac{hw}{k^2}=n$ , and only calculates the self-attention within each block:
\begin{equation}
\text{W-MSA}(\mbf{X})_i = \sm(Q_iK_i^\top/\sqrt{d})V_i.
    \label{eq:w-msa}
\end{equation}
As shown in \Cref{fig:attn}(c), the self-attention between blocks (i.e., along the $n$ axis) is overlooked. Fig.~\ref{fig:concerto}(b) provides another view of the resulting attention maps, where the attention outside the diagonal is omitted. W-MSA leaves out the relationships between blocks for efficiency. 

Another way to reduce computational cost is to use the transposed operation \cite{zamir2022restormer}, i.e., $\sm(Q^\top K)V^\top$, to reduce the computational complexity from $\mathcal{O}(h^2w^2)$ to $\mathcal{O}(hw)$. However, transposed self-attention lacks spatial awareness: random column permutations, or position swaps, do not make a difference to the resulting attention map, as shown in \Cref{fig:attn}(d).

To overcome the problems mentioned above, we propose the \textit{Concerto Self-Attention} (Section~\ref{sec:MA}).

\begin{figure*}[t]
\vspace{-5pt}
    \centering
    \includegraphics[width=.98\linewidth]{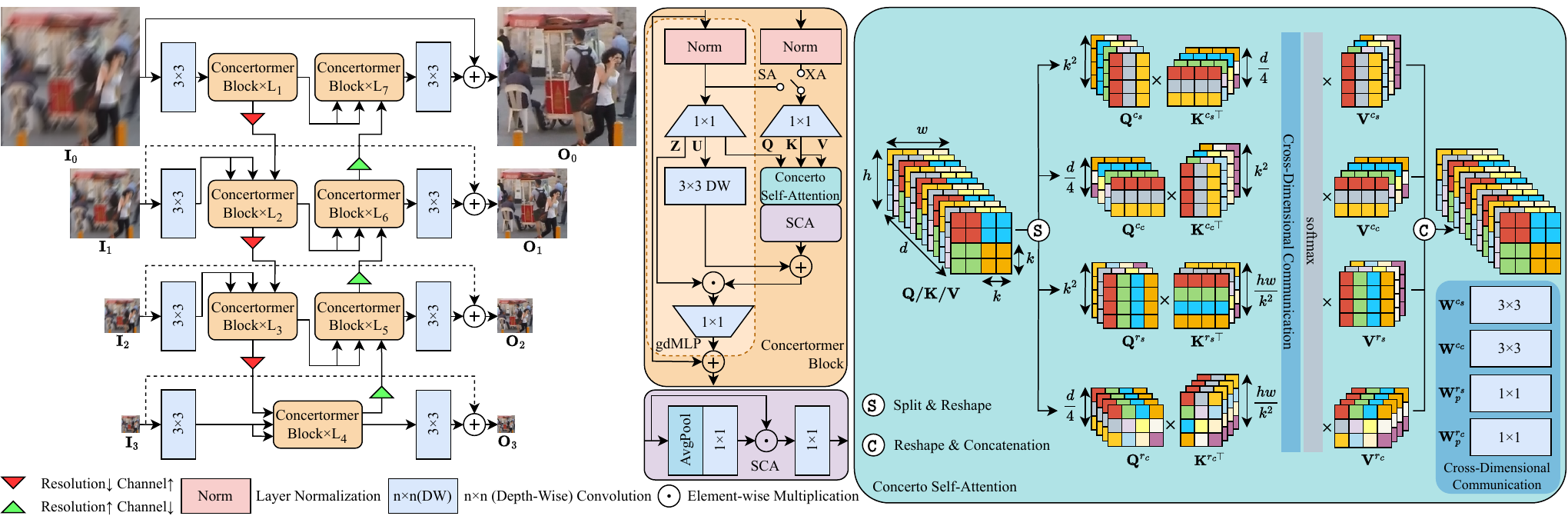}
        \vspace{-3mm}
    \caption{Network architecture. The overall network is shown on the left, and the sub-modules are on the right. XA: cross-attention, SA: self-attention. We use XA for the first blocks of $L_2 - L_7$, and SA for the remaining blocks.}
    \label{fig:model}
    \vspace{-6mm}
\end{figure*}

\section{Proposed Method}
\label{sec:method}
In this section, we describe the design methodology of the proposed {\ours}, whose building block is composed of a Concerto Self-Attention module and a gated-dconv MLP. We first give an overview (see Fig.~\ref{fig:model}), and then thoroughly describe the CSA module in Section~\ref{sec:MA}, the gdMLP in Section~\ref{sec:gdmlp}, and implementation details in Section~\ref{sec:detail}.

\ph{Overall Architecture.} Fig.~\ref{fig:model} shows the overall architecture on the left. We initially reduce its resolution by half using bilinear down-sampling to obtain $\mbf{I}_1 \in\mathbb{R}^{\frac{h}{2}\times \frac{w}{2}\times3}$, and similarly for $\mbf{I}_2\in\mathbb{R}^{\frac{h}{4}\times \frac{w}{4}\times3}$ and $\mbf{I}_3\in\mathbb{R}^{\frac{h}{8}\times \frac{w}{8}\times3}$. 

The original blurry image $\mbf{I}_0$ and its down-sampled versions $\mbf{I}_1 - \mbf{I}_3$ pass through a $3\times3$ convolution layer to increase channel dimensions to $1\times, 2\times, 3\times,$ and $4\times$ the base width, respectively, before being fed to the encoders. 
In the decoders, a $3\times3$ convolution layer with a skip connection to each input $\mbf{I}_0 - \mbf{I}_3$ produces the output images of different scales, used in the loss function (Section~\ref{sec:data}).
We replace traditional summation or concatenation between encoders and decoders with cross-attention (XA) on skip connections (see Section~\ref{sec:detail}), and lower-resolution inputs are similarly fed to the encoders via XA. 
Within the encoders, the resolution is halved and channels are doubled using $2\times2$ convolutions with stride 2, while in the decoders, resolution doubles, and channels halve using $1\times1$ convolutions followed by pixel-shuffle \cite{shi2016real}. 
Following \cite{zamir2022restormer, kong2023efficient}, block numbers increase from top to bottom in both encoders and decoders, with additional blocks allocated to the topmost decoder to refine the features before the final output.
\subsection{Concerto Self-Attention}
\label{sec:MA}

We propose a Concerto Self-Attention mechanism (see the right of Fig.~\ref{fig:model}) to reduce the time and space complexity of typical self-attention. 
The CSA prototype is defined as:
\begin{subequations}
 \label{eq:conerto_v1}
 \begin{align}
\text{CSA}(\mbf{X})_i &= (C_i + R)V_i,\label{eq:conerto_v1-1}\\
C_i &= \sm\left((Q_iK_i^\top - \overline{Q_iK_i^\top})/\alpha\right),\label{eq:conerto_v1-R}\\
R &= \sm(\overline{Q_iK_i^\top}/\beta),\label{eq:conerto_v1-C}
 \end{align}
\end{subequations}
where $\overline{Q_iK_i^\top}=\sum_{i=1}^n{Q_iK_i^\top}/n$, $\alpha$ and $\beta$ are learnable scalars. The \textit{Ripieno} component $R$ is applied to all $V_i$'s, computing the average attention across blocks to capture general spatial relationships, as shown in \eqref{eq:conerto_v1-C}. The \textit{Concertino} component $C_i$, specific to each $V_i$,  represents the difference from the average and gathers local information. We borrow the term ``concerto'' from music, where it refers to a composition featuring soloists (concertino) performing in dialogue with an accompanying orchestra (ripieno).

To compute the CSA~\eqref{eq:conerto_v1} efficiently, we divide the tensors into 2 parts, one for Concertino and one for Ripieno: we take $Q^c_i, Q^r_i, K^c_i, K^r_i, V^c_i, V^r_i \in \mathbb{R}^{k^2\times d_s/2}$ from $Q_i, K_i, V_i$, and compute the Concertino and Ripieno components as:
\begin{subequations}
 \label{eq:conerto_v2}
 \begin{align}
\text{CSA}(\mbf{X}) &= \begin{bmatrix}\mathcal{C}V^c &\mathcal{R}V^{r}\end{bmatrix},\label{eq:conerto_v2-1}\\
\mathcal{C} &= \diag(C_1,\dots,C_n), \label{eq:conerto_v2-RD}\\
\mathcal{R} &= \diag(R,\dots,R), \label{eq:conerto_v2-CD}\\
C_i &= \sm\left((Q^c_iK^{c\top}_i - \overline{Q^c_iK_i^{c\top}})/ \alpha\right),\label{eq:conerto_v2-R}\\
R &= \sm(\sum_iQ^r_iK^{r\top}_i/\beta),\label{eq:conerto_v2-C}
 \end{align}
\end{subequations}
where 
$V^c = \left[V^{c\top}_1 \dots V^{c\top}_n\right]^\top$, $V^r = \left[V^{r\top}_1 \dots V^{r\top}_n\right]^\top$ and $\diag(\cdot)$ denotes a block diagonal matrix. Since a linear projection layer generally follows the self-attention module, we assume that the addition in \eqref{eq:conerto_v1-1} can be absorbed into the projection layer and thus replaced with a concatenation operation in \eqref{eq:conerto_v2-1}, see Section~\ref{sec:CA} for further analysis. Additionally, by integrating the denominator $n$ into the learnable $\beta$, we use summation in \eqref{eq:conerto_v2-C} instead of average in \eqref{eq:conerto_v1-C}.

Note that \eqref{eq:conerto_v2-C} can be expressed as follows:
\begin{align}
\label{eq:qckc}
\sum_iQ^r_iK^{r\top}_i &= 
\begin{bmatrix}
Q^r_1 &\dots &Q^r_n
\end{bmatrix}
\begin{bmatrix}
K^{r\top}_1\\
\vdots\\
K^{r\top}_n
\end{bmatrix}=Q^rK^{r\top},
\end{align}
where $Q^r, K^r \in \mathbb{R}^{k^2\times \frac{d_sn}{2}}$. Thus, \eqref{eq:conerto_v2-C} can also be represented as a multiplication of two matrices.

\begin{figure}[tp]
    \centering
    \includegraphics[width=.8\linewidth]{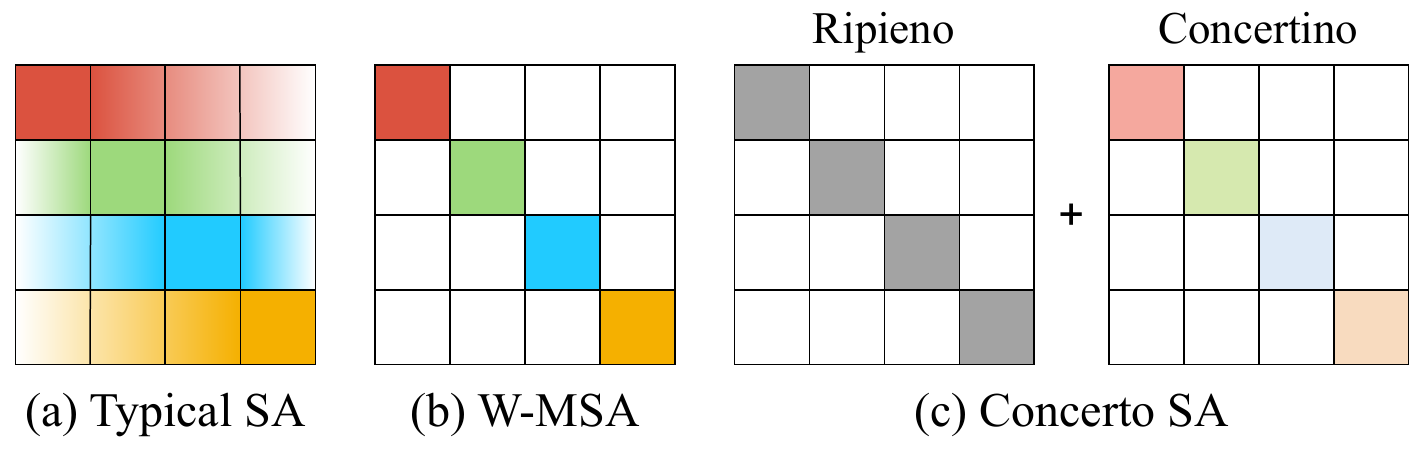}
     \vspace{-3mm}
    \caption{Attention maps. (a) Typical self-attention. The lighter color represents a smaller attention value. (b) Window multi-head self-attention. (c) Concerto Self-Attention, where each block on the diagonal shares the same Ripieno and has its own Concertino component.}
    \label{fig:concerto}
\end{figure}

Since the diagonal of \eqref{eq:conerto_v2-RD} can be regarded as a new dimension of the tensor, we express $\mathcal{C}$ in a tensor $\mbf{C}^s\in\mathbb{R}^{n\times k^2 \times k^2}$. To make the number of dimensions match, we choose $\frac{d_s}{2}$ as the number of additional heads in \eqref{eq:qckc}. Thus, we have the tensor version of $R$ as $\mbf{R}^s\in \mathbb{R}^{\frac{d_s}{2} \times k^2 \times k^2}$. Then \eqref{eq:conerto_v2-RD} and \eqref{eq:conerto_v2-C} can be expressed as
\begin{subequations}
\label{eq:conerto_tensor_s}
\begin{align}
&\mbf{C}^s = \sm\left(\mbf{Q}^{c_s}\times\mbf{K}^{c_s\top} - \overline{\mbf{Q}^{c_s}\times\mbf{K}^{c_s\top}}\right),\label{eq:conerto_tensor_rs} \\
&\mbf{R}^s = \sm\left(\mbf{Q}^{r_s}\times\mbf{K}^{r_s\top}\right),\label{eq:conerto_tensor_cs}
\end{align}
\end{subequations}
where $\mbf{Q}^{c_s}, \mbf{K}^{c_s}\in\mathbb{R}^{t\times n\times k^2 \times \frac{d_s}{2}}$, $\mbf{Q}^{r_s}, \mbf{K}^{r_s}\in\mathbb{R}^{t\times\frac{d_s}{2}\times k^2\times n}$, $t$ denotes the number of heads and $\times$ the matrix multiplication of the last two dimensions. For simplicity, we omit the learnable scalars $\alpha$ and $\beta$ here and in the remaining parts of this paper. \Cref{fig:concerto} illustrates the difference between CSA and typical SA and W-MSA in terms of attention maps. 

\ph{Cross-Dimensional Communication.}
In our CSA module, we have one more dimension, i.e., $n$ in \eqref{eq:conerto_tensor_rs} and $\frac{d_s}{2}$ in \eqref{eq:conerto_tensor_cs}, than the typical multi-head SA. Because the inner product is performed separately, we assume that building a connection between attention maps helps \cite{shazeer2020talking}. Thus, we propose the \textit{Cross-Dimensional Communication} (CDC) to utilize the information across dimensions.

For Ripieno tensor $\mbf{R}^s$, the $t \times \frac{d_s}{2}$ dimensions to be dealt with are constants, thus we use a $t \times \frac{d_s}{2}$ to $t \times \frac{d_s}{2}$ linear projection layer $\mbf{W}_p^{r_s}$; As for the Concertino tensor $\mbf{C}^s$, the dimension $n=\frac{hw}{k^2}$ is varying according to the input size, thus a convolution substitutes for linear projection: we reshape the logits in \eqref{eq:conerto_tensor_rs} to the size of $t\times \frac{h}{k}\times\frac{w}{k}\times k^4$, which is followed by a $3\times 3 \times1$ convolutional layer $\mbf{W}^{c_s}$, then reshape it back to the original shape. The Concerto Self-Attention with CDC can be expressed as
\begin{subequations}
\label{eq:cdc_s}
\begin{align}
\mbf{C}^s = \sm\left(\mbf{W}^{c_s}(\mbf{Q}^{c_s}\times\mbf{K}^{c_s\top})\right),\label{eq:cdc_rs} \\
\mbf{R}^s = \sm\left(\mbf{W}_p^{r_s}(\mbf{Q}^{r_s}\times\mbf{K}^{r_s\top})\right),\label{eq:cdc_cs}
\end{align}
\end{subequations}
where $\mbf{C}^{s}\in\mathbb{R}^{t\times n\times k^2 \times k^2}$, $\mbf{Q}^{c_s}, \mbf{K}^{c_s}\in\mathbb{R}^{t\times n\times k^2 \times \frac{d_s}{2}}$, $\mbf{R}^{s}\in\mathbb{R}^{t\times\frac{d_s}{2}\times k^2\times k^2}$ and $\mbf{Q}^{r_s}, \mbf{K}^{r_s}\in\mathbb{R}^{t\times\frac{d_s}{2}\times k^2\times n}$.
In \eqref{eq:cdc_rs}, we omit the mean $\overline{Q_i^cK_i^{c\top}}$ as it is a linear combination of $Q_i^cK_i^{c\top}$ and can be incorporated into $\mbf{W}^{c_s}$. Additionally, via a convolutional layer, the global mean $\overline{Q_i^cK_i^{c\top}}$ is replaced by a local mean $\sum_{j\subset\mathcal{N}(i)}Q_j^cK_j^{c\top}/|\mathcal{N}(i)|$, where $\mathcal{N}(i)$ denotes the blocks covered by the convolutional kernel centered at $i$. Since Ripieno $\mbf{R}^s$ already aggregates global information, the Concertino $\mbf{C}^s$ can focus on local details, making the use of a local mean over $\mathcal{N}(i)$ appropriate. See Section~\ref{sec:CA} for further analysis.

\ph{Channel CSA.} Although the transposed self-attention \cite{zamir2022restormer} is computationally efficient, it does not model the spatial information well as shown in Section~\ref{sec:revisit}.
To take advantage of complexity while ensuring that it can model spatial information well, we introduce the Concerto Self-Attention to the channel: taking $Q^c_i, Q^r_i, K^c_i, K^r_i, V^c_i, V^r_i \in \mathbb{R}^{d_c/2 \times k^2}$ from $Q_i, K_i, V_i$, we have similar results to \eqref{eq:conerto_v2}, \eqref{eq:qckc} and \eqref{eq:cdc_s}, except for $Q^r, K^r \in \mathbb{R}^{\frac{d_c}{2}\times k^2n}$, where we choose $k^2$ as the number of additional heads.
The channel Concerto Self-Attention can be obtained by:
\begin{subequations}
\label{eq:cdc_c}
\begin{align}
\mbf{C}^c = \sm\left(\mbf{W}^{c_c}(\mbf{Q}^{c_c}\times\mbf{K}^{c_c\top})\right),\label{eq:cdc_rc} \\
\mbf{R}^c = \sm\left(\mbf{W}_p^{r_c}(\mbf{Q}^{r_c}\times\mbf{K}^{r_c\top})\right),\label{eq:cdc_cc}
\end{align}
\end{subequations}
where $\mbf{C}^c\in \mathbb{R}^{t\times n\times \frac{d_c}{2}\times \frac{d_c}{2}}$ and $\mbf{R}^c\in \mathbb{R}^{t\times k^2\times \frac{d_c}{2}\times \frac{d_c}{2}}$. Since the position is separately recorded in the $n$ dimension of $\mbf{C}^c$ and $k^2$ of $\mbf{R}^c$, our channel Concerto Self-Attention can sense spatial information.

\ph{Mixing Projection.} Applying the Concerto Self-Attention to both spatial and channel dimension, we have $\mbf{X}^A = \mathtt{C}(\mbf{C}^s\times\mbf{V}^{c_s},\mbf{R}^s\times\mbf{V}^{r_s},\mbf{C}^c\times\mbf{V}^{c_c},\mbf{R}^c\times\mbf{V}^{r_c})$, where $\mbf{V}^{c_s}\in\mathbb{R}^{t\times n\times k^2 \times \frac{d_s}{2}}$, $\mbf{V}^{r_s}\in\mathbb{R}^{t\times\frac{d_s}{2}\times k^2\times n}$, $\mbf{V}^{c_c}\in\mathbb{R}^{t\times n\times \frac{d_c}{2} \times k^2 }$ and $\mbf{V}^{r_c}\in\mathbb{R}^{t\times k^2\times \frac{d_c}{2}\times n}$, and $\mathtt{C}(\cdot)$ reshapes then concatenates tensors to $\mathbb{R}^{h\times w\times d}$. 

Although these four attentions compute the spatial and channel relationships exhaustively, they are operated independently. Thus, we also use simplified channel attention (SCA) \cite{chen2022simple} to balance their importance, which is defined as
\begin{equation}
    \label{eq:sca}
    SCA(\mbf{X}^A) = \mbf{W}_p^2(\mbf{W}_p^1(AvgPool(\mbf{X}^A)) \odot_c \mbf{X}^A),
\end{equation}
where $\odot_c$ denotes the channel-wise multiplication \cite{hu2018squeeze}, $AvgPool$ the global average pooling operation \cite{lin2013network}, and $\mbf{W}_p^1$ and $\mbf{W}_p^2$ are the functions of $1\times 1$ convolution. \eqref{eq:sca} is the final output of the Concerto Self-Attention module.


\subsection{Gated-Dconv MLP}
\label{sec:gdmlp}

To reduce the complexity of the original Transformer FFN and enhance feature representation for image restoration, we propose the gated-dconv MLP, which works with the self-attention module within a single block (see the middle box of Fig.~\ref{fig:model}). The gdMLP can be expressed as
\begin{subequations}
\label{eq:gdmlp}
\begin{align}
    gdMLP(\mbf{X}) = \mbf{W}_p^g(\mbf{U} \odot \mbf{Z}), \label{eq:gdmlpout} \\
    \mbf{U} =  \mbf{W}_d^u(\mbf{W}_p^u(\mbf{X})), \\
    \mbf{Z} =  \mbf{W}_p^z\mbf{X}, \label{eq:z}
\end{align}
\end{subequations}
where $\odot$ denotes the element-wise product, and $\mbf{W}_d^u$ is the weight matrix of $3\times3$ dconv; $\mbf{W}_p^u$, $\mbf{W}_p^z$ and $\mbf{W}_p^g$ are the weight matrices of $1\times 1$ convolution, respectively. To replace FFN, gdMLP has to incorporate the expansion-and-compression operation, so $\mbf{W}_p^z$ and $\mbf{W}_p^u$ double the channels and $\mbf{W}_p^g$ half the channels.

Unlike the gated MLP \cite{liu2021pay}, we omit layer normalization and replace the spatial projection (a large linear layer for deblurring) with depth-wise convolution. This adjustment makes the module fully convolutional, which is efficiently suited for image restoration tasks.

To cooperate with the Concerto Self-Attention, as shown in Fig.~\ref{fig:model}, we modify \eqref{eq:gdmlpout} as:
\begin{equation}
    \label{eq:Concertormer}
    gdMLP(\mbf{X}) = \mbf{W}_p^g( (SCA(\mbf{X}^A) + \mbf{U})  \odot \mbf{Z}).
\end{equation}
With the depth-wisely convolved tensor $\mbf{U}$, the discontinuous border of non-overlapping blocks would be compensated.

\subsection{Implementation Details}
\label{sec:detail}
We choose the channel size $d_s=d_c=\frac{d}{2t}$, and block size $k=8$. 
In the whole model, i.e., the left of Fig.~\ref{fig:model}, we apply the cross-attention mechanism to the first block in each level except for the topmost encoder. For the cross-attention blocks, whose $\mbf{K}$ and $\mbf{V}$ in \eqref{eq:qkv} would be $ K = \mbf{W}_p^k(\mbf{Y})$ and $V = \mbf{W}_p^v(\mbf{Y})$, where $\mbf{Y}$ is the input other than $\mbf{X}$. As shown in Fig.~\ref{fig:model}, the $\mbf{Y}$ of encoders comes from down-scaled inputs followed by a $3\times3$ convolution; for decoders, $\mbf{Y}$ is the output of the encoder of the corresponding level.

We adopt multi-head attention in our models, where the number of heads increases with the channel number: the top encoder and decoder have a single head, and the layers beneath have double the heads of the above level.

{\ours} and {\ourslite} both have channel width $48$, which is the output channel size of the top left $3\times3$ convolution on the left of \Cref{fig:model}, and the expansion factor of gdMLP is set as $2$, i.e., the channel numbers of $\mbf{W}^z_p, \mbf{W}^u_p$ in \eqref{eq:gdmlp} and $\mbf{W}^q_p, \mbf{W}^k_p, \mbf{W}^v_p$ in \eqref{eq:qkv} double the channels. The $L_1$ to $L_7$, i.e., the block numbers shown in \Cref{fig:model}, of {\ourslite} are $[4, 4, 12, 2, 12, 4, 4]$ and the full {\ours} $[6, 8, 24, 2, 24, 8, 8]$.

\section{Experimental Results}
\label{sec:result}
In this section, we first describe the datasets and training settings and compare our method with state-of-the-art ones in image deblurring. In addition, we show that our method can be applied to image deraining and real image denoising.

\vspace{-4pt}
\subsection{Datasets and Training Details}
\label{sec:data}
\vspace{-4pt}
Similarly to existing methods~\cite{nah2017deep,zamir2021multi,cho2021rethinking}, we train our {\ours} on the \textsc{GoPro} dataset, and evaluate it on \textsc{GoPro} \cite{nah2017deep} and \textsc{HIDE} \cite{shen2019human} for fairness. 
We also evaluate our method on the real-world deblurring datasets, i.e., \textsc{Realblur-R} and \textsc{Realblur-J} by \cite{rim2020real}. 
We use the model trained on \textsc{GoPro} and fine-tune it using \cite{rim2020real}. 

For simplicity, we illustrate the training process for image deblurring on \textsc{GoPro}; other tasks are detailed in the supplementary materials. We use $\ell_1$ loss in the spatial domain and frequency domain as \cite{cho2021rethinking}, where the ground-truth and its 3 down-sampled duplicates are used to calculate the loss between outputs $\mbf{O}_0-\mbf{O}_3$ in Fig.~\ref{fig:model}. As for training settings, we use AdamW \cite{loshchilov2017decoupled} with $\beta_1=\beta_2=0.9$, and weight decay of $10^{-3}$. Progressive training begins with $128\times128$ patches and a batch size of 64, then scales to $256\times256$ with batch size 16, and $320\times320$ with batch size 8, each for 200,000 iterations. The learning rate starts at $10^{-3}$, decaying to $10^{-7}$ via Cosine Annealing.
The training patches are randomly cropped from $512\times 512$ overlapped patches and augmented by random flips and $90^{\circ}$ rotations.

In addition to regular inference, we apply Test-time Local Converter (TLC) \cite{chu2022improving} during inference. The better results among the TLC inference and the regular inference are shown in this section.
\begin{table}[tp]
    \centering
\caption{Results on the \dataset{GoPro} \cite{nah2017deep} and \dataset{HIDE} \cite{shen2019human} dataset, where our two models achieve the \fst{best} and the \snd{second-best} results. }
\vspace{-3mm}
\scriptsize
    \begin{tabular}{rcccc}
    \toprule
\mr{2}{Method} &\mc{2}{\dataset{GoPro}} &\mc{2}{\dataset{HIDE}}\\
&PSNR &SSIM &PSNR &SSIM \\ \midrule
DeblurGAN-v2 \cite{kupyn2019deblurgan}	 &29.55	&0.934	&26.61	&0.875	\\
SRN \cite{tao2018scale}	                 &30.26	&0.934	&28.36	&0.904	\\
DMPHN \cite{zhang2019deep}	             &31.20	&0.945	&29.09	&0.924	\\
SAPHN \cite{suin2020spatially}	         &31.85	&0.948	&29.98	&0.930	\\
MIMO-UNet+ \cite{cho2021rethinking}	     &32.45	&0.957	&29.99	&0.930	\\
MPRNet \cite{zamir2021multi}	         &32.66	&0.959	&30.96	&0.939	\\
DeepRFT+ \cite{mao2021deep}              &32.23 &0.963  &31.42  &0.944  \\
Restormer \cite{zamir2022restormer} 	 &32.92	&0.961	&31.22	&0.942	\\
Uformer \cite{wang2022uformer}           &33.06 &0.967  &30.90  &0.953  \\
Stripformer \cite{tsai2022stripformer}	 &33.08	&0.962	&31.03	&0.940	\\
MPRNet-local \cite{chu2022improving}     &33.31 &0.964  &31.19  &0.942  \\
Restormer-local \cite{chu2022improving}	 &33.57	&0.966	&31.49	&0.945	\\
NAFNet \cite{chen2022simple}	         &33.71	&0.967	&31.32	&0.943	\\
GRL	\cite{li2023efficient}               &33.93	&0.968	&31.65	&0.947	\\
FFTformer \cite{kong2023efficient}	     &\snd{34.21}	&0.969	&31.62	&0.946	\\
DiffIR \cite{xia2023diffir}	             &33.31	&0.964	&31.50	&0.946	\\ 
MLWNet \cite{gao2024efficient}           &33.86 &0.968  &31.36  &0.945 \\
MISCFilter \cite{liu2024motion}           &34.10 &0.969  &31.66  &0.946 \\
\midrule
{\ourslite}	             &34.18	&\snd{0.970} &\snd{31.82} &\snd{0.949}\\
{\ours}	                 &\fst{34.42} &\fst{0.971} &\fst{32.12} &\fst{0.951}\\ \bottomrule
    \end{tabular}
    \label{tab:mdb}
\end{table}
\begin{table}[tp]
    \caption{Results on the \dataset{RealBlur} dataset~\cite{rim2020real}.}
    \centering
 \vspace{-3mm}
\scriptsize
    \begin{tabular}{rcccc}
    \toprule
\mr{2}{Method} &\mc{2}{\dataset{RealBlur-R}} &\mc{2}{\dataset{RealBlur-J}}\\
&PSNR &SSIM &PSNR &SSIM \\ \midrule
DeblurGAN-v2 \cite{kupyn2019deblurgan}        &35.11  &0.935  &28.69  &0.866 \\
SRN \cite{tao2018scale} &38.65  &0.965  &31.38  &0.909 \\
MIMO-UNet+ \cite{cho2021rethinking}         &-  &-  &31.92  &0.919 \\
MAXIM-3S \cite{tu2022maxim}  &39.45 &0.962 &32.84  &0.935 \\
BANet \cite{tsai2022banet}         &39.55 &0.971 &32.00 &0.923 \\
DeepRFT+ \cite{mao2021deep}         &39.70 &0.971  &32.18 &0.928 \\
Stripformer \cite{tsai2022stripformer}         &39.84  &0.974  &32.48  &0.929 \\
FMIMO-UNet+ \cite{mao2023intriguing} &39.98 &0.973 &32.56 &0.932 \\
FFTformer \cite{kong2023efficient}  &40.11  &0.973  &32.62  &0.933 \\
GRL	\cite{li2023efficient}  &40.20 &0.974 &32.82 &0.932 \\
MLWNet \cite{gao2024efficient}         &\snd{40.69}        &\snd{0.976}  &\snd{33.84}  &\snd{0.941} \\
MISCFilter \cite{liu2024motion}      &40.61\footnotemark[1] &0.975\tablefootnote{These numbers are different from those reported in their paper, while we have confirmed these are correct with the authors.}       &\fst{33.88}  &0.938 \\
\midrule
{\ours}         &\fst{40.78}  &\fst{0.977}  &33.51  &\fst{0.945} \\\bottomrule
    \end{tabular}
    \label{tab:realblur}
\end{table}
\begin{figure*}
\sbox{\measurebox}{
\begin{minipage}[b]{.34\linewidth} 
\subfloat[Blurry Input]{\includegraphics[width=1.\linewidth]{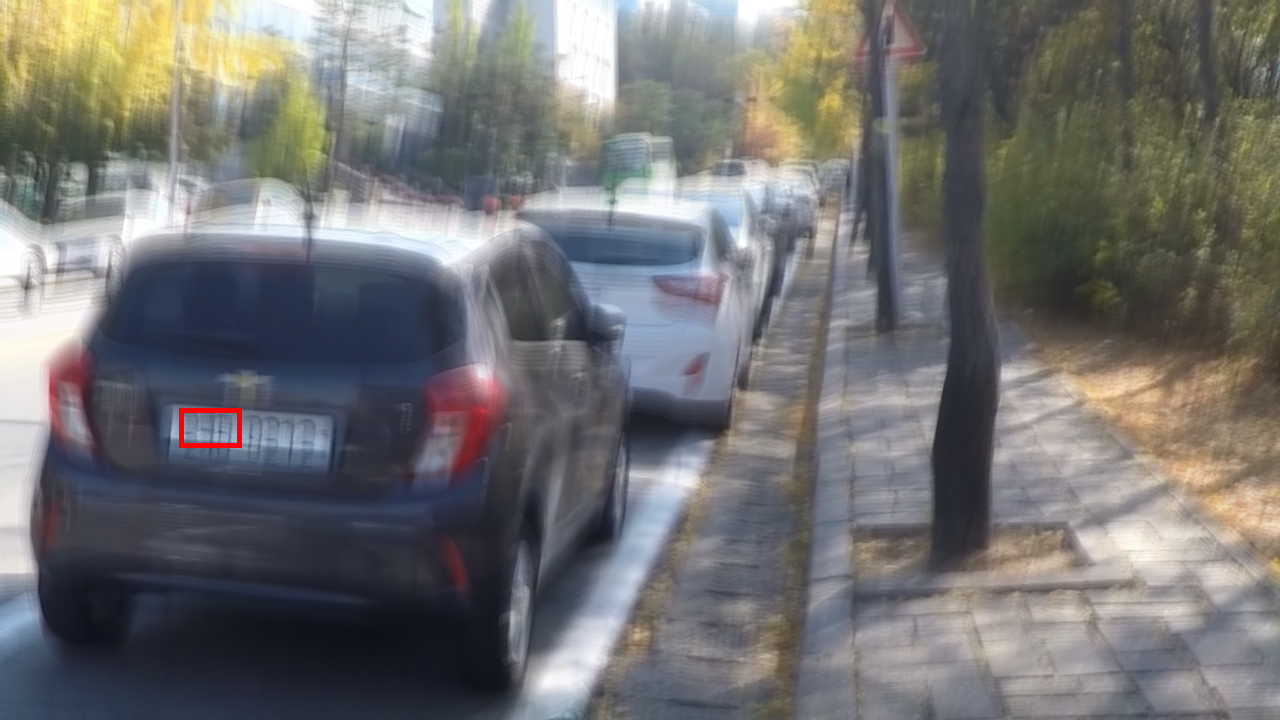}}
\end{minipage}}\hfill
\usebox{\measurebox}
\begin{minipage}[b][\ht\measurebox][s]{.64\linewidth} 
\subfloat[MPRNet]{\includegraphics[width=0.195\linewidth]{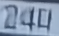}}\hfill
\subfloat[Restormer]{\includegraphics[width=0.195\linewidth]{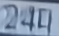}}\hfill
\subfloat[Uformer]{\includegraphics[width=0.195\linewidth]{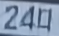}}\hfill
\subfloat[Stripformer]{\includegraphics[width=0.195\linewidth]{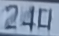}}\hfill
\subfloat[NAFNet]{\includegraphics[width=0.195\linewidth]{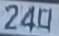}}\\  \vfill
\subfloat[GRL]{\includegraphics[width=0.195\linewidth]{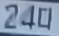}}\hfill
\subfloat[FFTformer]{\includegraphics[width=0.195\linewidth]{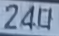}}\hfill
\subfloat[DiffIR]{\includegraphics[width=0.195\linewidth]{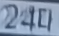}}\hfill
\subfloat[\ours]{\includegraphics[width=0.195\linewidth]{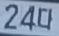}}\hfill
\subfloat[GT]{\includegraphics[width=0.195\linewidth]{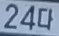}}
\end{minipage}
\vspace{-3mm}
      \caption{Single-image motion deblurring on the \textsc{GoPro} \cite{nah2017deep} dataset.}
\label{fig:gopro}
\vspace{-3mm}
\end{figure*}
\begin{figure*}
\sbox{\measurebox}{
\begin{minipage}[b]{.24\linewidth} 
\subfloat[Blurry Input]{\includegraphics[width=1.\linewidth]{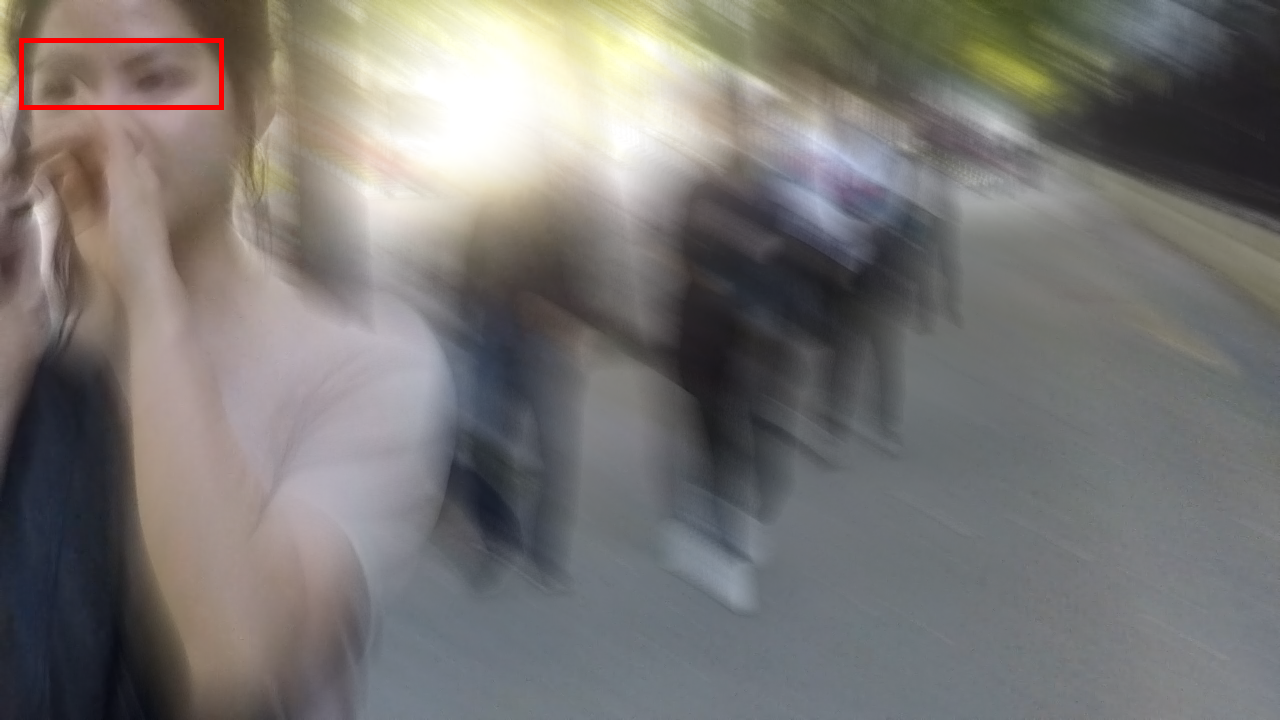}}
\end{minipage}}
\usebox{\measurebox}
\begin{minipage}[b][\ht\measurebox][s]{.74\linewidth} 
\subfloat[MPRNet]{\includegraphics[width=0.195\linewidth]{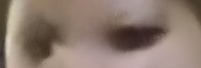}}\hfill
\subfloat[Restormer]{\includegraphics[width=0.195\linewidth]{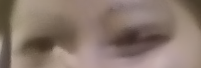}}\hfill
\subfloat[Uformer]{\includegraphics[width=0.195\linewidth]{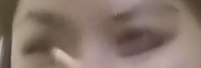}}\hfill
\subfloat[Stripformer]{\includegraphics[width=0.195\linewidth]{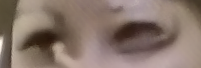}}\hfill
\subfloat[NAFNet]{\includegraphics[width=0.195\linewidth]{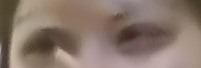}}\\ \vfill
\subfloat[GRL]{\includegraphics[width=0.195\linewidth]{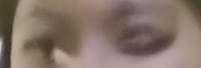}}\hfill
\subfloat[FFTformer]{\includegraphics[width=0.195\linewidth]{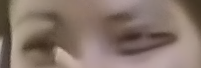}}\hfill
\subfloat[DiffIR]{\includegraphics[width=0.195\linewidth]{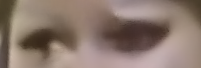}}\hfill
\subfloat[\ours]{\includegraphics[width=0.195\linewidth]{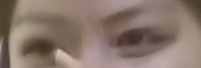}}\hfill
\subfloat[GT]{\includegraphics[width=0.195\linewidth]{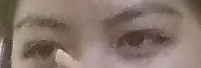}}
\end{minipage}
\vspace{-3mm}
      \caption{Single-image motion deblurring on the \textsc{HIDE} \cite{shen2019human} dataset.}
\label{fig:hide}
\vspace{-4mm}
\end{figure*}

\subsection{Comparisons with State-of-the-Arts}
\label{sec:deblur}
%
%
\vspace{-3pt}
\ph{Evaluations on Synthetic Datasets.}
Table~\ref{tab:mdb} shows comparisons on the synthetic datasets \textsc{GoPro} \cite{nah2017deep} and \textsc{HIDE} \cite{shen2019human}. Our models achieve the best and second-best results on these datasets. Compared to the FFTformer \cite{kong2023efficient}, our {\ours} further boosts the performance by $0.21$ dB in terms of PSNR. 
Our lite model uses $16\%$ fewer FLOPs while maintaining competitive performance, demonstrating the efficiency and effectiveness of {\ours} (Fig.~\ref{fig:flops}).

Fig.~\ref{fig:gopro} shows the visual comparisons on \textsc{GoPro}: our model generates the fewest artifacts and the sharpest characters among these methods. We demonstrate the qualitative evaluation on \textsc{HIDE} in Fig.~\ref{fig:hide}, where only our approach restores the eyes.

\begin{table}[tp]
    \centering
    \scriptsize
    \caption{Image deblurring with JPEG artifacts.}
     \vspace{-3mm}
\tabcolsep=2pt
    \begin{tabular}{ccccccc}\toprule
\mr{2}{Dataset} &\mr{2}{Metrics}  &MPRNet  &HINet  &MAXIM &NAFNet  &\mr{2}{\ours} \\
& &\cite{zamir2021multi} &\cite{chen2021hinet}  &\cite{tu2022maxim} &\cite{chen2022simple} &\\\midrule
REDS-val-300        &PSNR  &28.79 &28.83  &28.93  &\snd{29.09} &\fst{29.33}\\
\cite{nah2021ntire} &SSIM  &0.811 &0.862  &0.865  &\snd{0.867} &\fst{0.872} \\ \bottomrule
\end{tabular}
        \label{tab:jdb}
\end{table}
\ph{Evaluations on Real-World Datasets.}
Table~\ref{tab:realblur} shows the deblurring results on the \dataset{RealBlur} dataset. We compare our method to models trained or fine-tuned on \dataset{RealBlur}, where our methods generate the highest PSNR and SSIM. 

\begin{table}[tp]
    \centering
    \scriptsize
    \caption{Image deraining results. We test on \dataset{Test100} \cite{zhang2019image}, \dataset{Rain100H} \cite{yang2017deep}, \dataset{Rain100L} \cite{yang2017deep}, \dataset{Test2800} \cite{fu2017removing}.}
    \vspace{-3mm}
\rowcolors{2}{gray!20}{white}
\tabcolsep=1pt
    \begin{tabular}{rcccccccccc}
    \toprule
         &\mc{2}{\dataset{Test100}} &\mc{2}{\dataset{Rain100H}} &\mc{2}{\dataset{Rain100L}}  &\mc{2}{\dataset{Test2800}}  &\mc{2}{Average}\\
  \mr{-2}{Method}  &PSNR &SSIM  &PSNR &SSIM  &PSNR &SSIM   &PSNR &SSIM  &PSNR &SSIM\\ \midrule
DerainNet
&22.77 &0.810 &14.92 &0.592 &27.03 &0.884 &24.31 &0.861  &22.26	 &0.787\\ 
SEMI
&22.35 &0.788 &16.56 &0.486 &25.03 &0.842 &24.43 &0.782  &22.09	 &0.725\\ 
DIDMDN
&22.56 &0.818 &17.35 &0.524 &25.23 &0.741 &28.13 &0.867  &23.32	 &0.738\\ 
UMRL
&24.41 &0.829 &26.01 &0.832 &29.18 &0.923 &29.97 &0.905  &27.39	 &0.872\\ 
RESCAN   
&25.00 &0.835 &26.36 &0.786 &29.80 &0.881 &31.29 &0.904  &28.11	 &0.852\\ 
PreNet 
&24.81 &0.851 &26.77 &0.858 &32.44 &0.950 &31.75 &0.916  &28.94	 &0.894\\ 
MSPFN
&27.50 &0.876 &28.66 &0.860 &32.40 &0.933 &32.82 &0.930  &30.35	 &0.900\\ 
MPRNet
&30.27 &0.897 &30.41 &0.890 &36.40 &0.965 &33.64 &0.938  &32.68	 &0.923\\ 
SPAIR 
&30.35 &0.909 &30.95 &0.892 &36.93 &0.969 &33.34 &0.936  &32.89	 &0.927\\
Restormer
&\snd{32.00} &\snd{0.923} &\snd{31.46} &\snd{0.904} &\snd{38.99} &\snd{0.978} &\snd{34.18} &\snd{0.944} &\snd{34.16} &\snd{0.937}\\\midrule
{\ours} &\fst{32.19} &\fst{0.926} &\fst{32.71} &\fst{0.921} &\fst{39.36} &\fst{0.978} &\fst{34.21} &\fst{0.946} &\fst{34.62} &\fst{0.943}\\ \bottomrule 
    \end{tabular}
    \label{tab:drain}
\end{table}

\vspace{-2mm}
\subsection{More Applications}
\vspace{-1mm}
\label{sec:drain}
We further show that our method can be applied to other related image restoration problems.

\ph{Image Deblurring with JPEG Artifacts.}
We train the model on \dataset{REDS} \cite{nah2021ntire} and test on \dataset{REDS-val-300} \cite{chen2021hinet, tu2022maxim}. 
Table~\ref{tab:jdb} shows that 
{\ours} further boosts the performance by more than $0.2$ dB in PSNR.

\begin{figure*}
\sbox{\measurebox}{
\begin{minipage}[b]{.14\linewidth} 
\subfloat[Rainy Input]{\includegraphics[width=1.\linewidth]{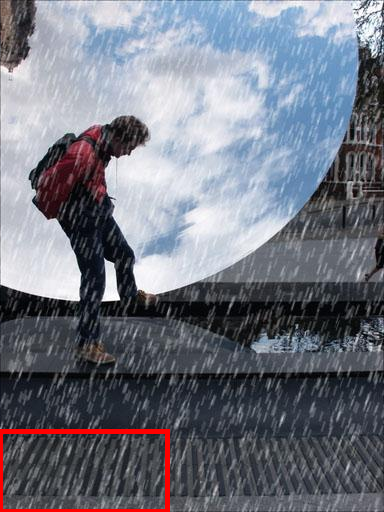}}
\end{minipage}}
\usebox{\measurebox}
\begin{minipage}[b][\ht\measurebox][s]{.84\linewidth} 
\subfloat[DerainNet]{\includegraphics[width=0.194\linewidth]{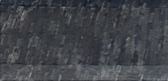}}\hfill
\subfloat[SEMI]{\includegraphics[width=0.194\linewidth]{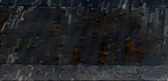}}\hfill
\subfloat[DIDMDN]{\includegraphics[width=0.194\linewidth]{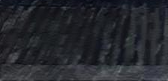}}\hfill
\subfloat[UMRL]{\includegraphics[width=0.194\linewidth]{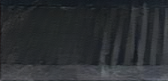}}\hfill
\subfloat[RESCAN]{\includegraphics[width=0.194\linewidth]{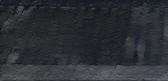}}\\ \vfill
\subfloat[PreNet]{\includegraphics[width=0.194\linewidth]{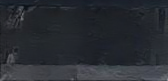}}\hfill
\subfloat[MPRNet]{\includegraphics[width=0.194\linewidth]{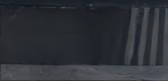}}\hfill
\subfloat[Restormer]{\includegraphics[width=0.194\linewidth]{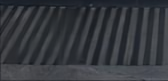}}\hfill
\subfloat[\ours]{\includegraphics[width=0.194\linewidth]{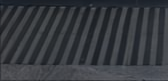}}\hfill
\subfloat[GT]{\includegraphics[width=0.194\linewidth]{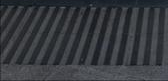}}
\end{minipage}
\vspace{-3mm}
      \caption{Deraining on the \dataset{Test100} \cite{zhang2019image} dataset.}
\label{fig:derain}
\vspace{-4mm}
\end{figure*}

\ph{Image Deraining.}
Similar to~\cite{zamir2022restormer}, we train our model on \textsc{Rain13K} \cite{zamir2021multi} and test it on \textsc{Rain100H} \cite{xu2013unnatural} and \textsc{Rain100L} \cite{xu2013unnatural}, \textsc{Test100} \cite{zhang2019image}, and \textsc{Test2800} \cite{fu2017removing}. 
Following \cite{jiang2020multi, purohit2021spatially, zamir2021multi, zamir2022restormer}, we compute evaluation metrics with Y channel of the YCbCr color space. Table~\ref{tab:drain} shows that our method performs better on image deraining. Fig.~\ref{fig:derain} shows that {\ours} restores the images with better textures.

\vspace{-3pt}
\section{Analysis and Discussion}
We have shown the effectiveness and efficiency of our method. This section analyzes the main components of {\ours} and the Concerto Self-Attention. The models in this section are constructed according to {\ourslite} model.
Patches of $256\times 256$ pixels of \dataset{GoPro} dataset and batch size $8$ are used for 200,000 iterations, and the learning rate decreases from $10^{-3}$ to $10^{-6}$ by Cosine Annealing. FLOPs are computed via a $256\times256$ input image. The other settings follow the formal {\ours} models described in Section \ref{sec:data}.

\subsection{Ablation Study}
\label{sec:abl}
\begin{table}
\scriptsize
    \centering
    \caption{Ablation study on the \textsc{GoPro} dataset. gdMLP: gated-dconv MLP, FFN: Feed-Forward Network, SCA: simplified channel attention, MH: multi-head, CDC: Cross-Dimensional Communication, $\mbf{R}^s$: spatial Ripieno $\mbf{C}^s$: spatial Concertino, $\mbf{R}^c$: channel Ripieno, $\mbf{C}^c$: channel Concertino. The FLOPs are presented in Giga, and the parameters in Mega.}
\vspace{-3mm}
\resizebox{\linewidth}{!}{
\tabcolsep=2pt
    \begin{tabular}{cccccccccccccc}
    \toprule
        &\mc{2}{Architecture} &\mc{4}{Concerto SA}  & & & &\mc{2}{Model Size}  &\mc{2}{Metrics}\\ 
Model  &FFN &gdMLP &$\mbf{C}^s$ &$\mbf{R}^s$ &$\mbf{C}^c$ &$\mbf{R}^c$ &\mr{-2}{SCA} &\mr{-2}{MH} & \mr{-2}{CDC}   &FLOPs &Param &PSNR  &SSIM\\ \midrule
0     &\checkmark & & & & & & & & &52.87 &11.0 &26.28 &0.851 \\
1&    &\checkmark & & & & & & & &41.22 &8.5 &32.35 &0.951 \\ \cmidrule{11-12}
2&    &\checkmark &\checkmark & & & & & & & & &32.58 &0.953 \\ 
3&    &\checkmark &\checkmark &\checkmark & & & & & &\mr{-2}{119.34} &  &33.11 &0.958 \\ \cmidrule{11-11}
4&    &\checkmark & & &\checkmark & & & & &155.38 & &32.65 &0.953 \\ 
5&    &\checkmark & & &\checkmark &\checkmark & & & &130.01 & &33.00 &0.956 \\ 
6&    &\checkmark &\checkmark &\checkmark &\checkmark &\checkmark & & & &118.33 &\mr{-6}{21.2}  &33.20 &0.958 \\ \cmidrule{12-12} 
7&    &\checkmark &\checkmark &\checkmark &\checkmark &\checkmark &\checkmark & & &118.57 &\mr{2}{26.3} &33.31 &0.959 \\ 
8&    &\checkmark& &\checkmark &\checkmark &\checkmark &\checkmark &\checkmark & &114.98 & &33.35 &0.960 \\   \cmidrule{12-12}
9&    &\checkmark &\checkmark &\checkmark &\checkmark &\checkmark &\checkmark &\checkmark &\checkmark &116.79 &28.9 &\fst{33.53} &\fst{0.961} \\  
10    &\checkmark &&\checkmark &\checkmark &\checkmark &\checkmark &\checkmark &\checkmark &\checkmark &116.81 &29.2 &31.90 &0.945 \\\bottomrule
    \end{tabular}}
    \label{tab:abl}
\end{table}

To show the effect of each component in \Cref{tab:abl}, we modify the building blocks to examine the proposed method. These models are numbered for easier and clearer discussion:
The study starts from a typical MLP (model 0); the Baseline (model 1) is the vanilla gdMLP without self-attention, and then we gradually add Concerto Self-Attention to gdMLP: the model 2 contains the self-attention mechanism of \eqref{eq:cdc_rs} without $\mbf{W}^{r_s}$; the model 3 contains \eqref{eq:cdc_s} without $\mbf{W}$'s; the model 4 uses \eqref{eq:cdc_rc} without $\mbf{W}^{r_c}$ and model 5 uses \eqref{eq:cdc_c} without $\mbf{W}$'s; model 6 uses the full CSA mechanism. Model 7 tests the effect of SCA \eqref{eq:sca}, model 8 for multi-head attention, and model 9 is the complete {\ours}.

The gdMLP architecture improves PSNR by over $6$ dB while reducing FLOPs and parameters by more than $22\%$ compared to the MLP. From model 2 to model 3, spatial CSA increases PSNR by over $0.5$ dB, while channel CSA shows a $0.35$ dB improvement (model 4 to model 5). For model 6, combining spatial and channel CSA boosts PSNR by about $0.85$ dB over the Baseline. Model 7’s SCA adds $0.11$ dB PSNR with simply $0.2\%$ additional FLOPs. Finally, CDC enhances PSNR by $0.18$ dB from model 8 to model 9, with a cost of $1.57\%$ FLOPs and $9.89\%$ parameters. It is worth noting that dividing channels for multiple self-attention mechanisms reduces FLOPs since the vector lengths for inner products are shortened, thereby the FLOPs decrease from model 4 to model 6, and the same trend can also be observed with multi-head attention, i.e., model 7 to model 8. This further reduces the computational complexity of CSA.

To validate the effectiveness of the single-stage gdMLP architecture, we train a two-staged model combining Concerto Self-Attention and an FFN (model 10 in Table~\ref{tab:abl}). With an adjusted FFN expansion factor to match the FLOPs and parameters of {\ours}, our method achieves higher PSNR by more than $1.6$ dB.
\vspace{-5pt}
\subsection{Analysis of Concerto Self-Attention}
\label{sec:CA}
\begin{figure}
    \centering
    \subfloat[\label{fig:lama}]{\includegraphics[width=0.14\linewidth]{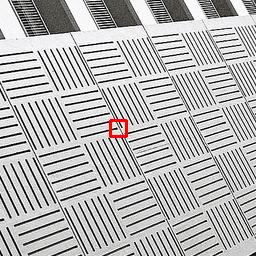}}\hfill
    \subfloat[\label{fig:lamb}]{\includegraphics[width=0.14\linewidth]{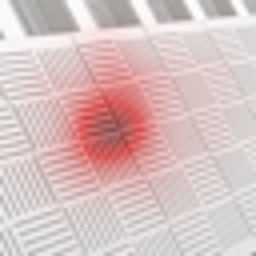}}\hfill
    \subfloat[\label{fig:lamc}]{\includegraphics[width=0.14\linewidth]{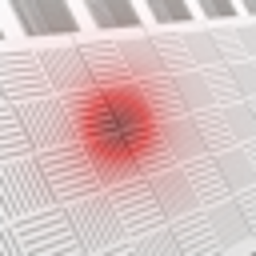}}\hfill
    \subfloat[\label{fig:lamd}]{\includegraphics[width=0.14\linewidth]{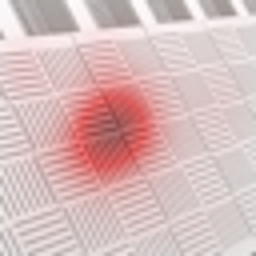}}\hfill
    \subfloat[\label{fig:lame}]{\includegraphics[width=0.14\linewidth]{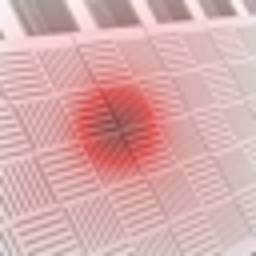}}\hfill
    \subfloat[\label{fig:lamf}]{\includegraphics[width=0.14\linewidth]{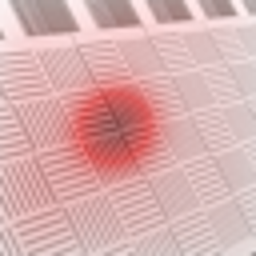}}\hfill
    \subfloat[\label{fig:lamg}]{\includegraphics[width=0.14\linewidth]{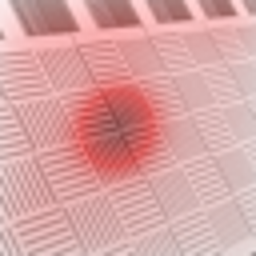}}
    \vspace{-3mm}
    \caption{Comparison of local attribution maps \cite{gu2021interpreting}. (a) Ground truth and target patch. (b)  - (g) Contribution areas of spatial Concertino without and with shifted windows, Prototype, Concerto Self-Attention, and CDC with and without global mean, respectively.}
    \label{fig:lam}
\end{figure}

\vspace{-3pt}
We have presented the derivation of our method in Section~\ref{sec:MA}. Here, we analyze how each step contributes to the satisfactory results compared to state-of-the-art works.  The diffusion index (DI) in \Cref{tab:concerto} and the local attribution maps (LAMs) in \Cref{fig:lam} are used to quantify and visualize the receptive field \cite{gu2021interpreting}, respectively.

In \Cref{tab:concerto}, we analyze the models from \eqref{eq:conerto_v1} to \eqref{eq:cdc_c}. Comparing the Prototype to the spatial Concertino (model 2 of able~\ref{tab:abl}), although the PSNR and SSIM slightly decrease, the DI increases for about $32\%$.  This suggests that the average operation in \eqref{eq:conerto_v1} significantly expands the receptive field with negligibly additional computations, and achieves higher DI than spatial Concertino with shifted windows.
\cref{fig:lamb}-\labelcref{fig:lamd} show that the Prototype model extends self-attention to distant and correlated regions, such as the top-left corner. Using concatenation in \eqref{eq:conerto_v2} (Concerto) instead of addition in \eqref{eq:conerto_v1} (Prototype) results in higher PSNR, SSIM, and DI, supporting the assumption that linear projection can replace addition.
The bottom row of Table~\ref{tab:concerto} shows the effect of the proposed CDC, which has the highest PSNR, SSIM, and DI at the cost of simply $0.36\%$ more FLOPs and $6.13\%$ more parameters. The largest receptive field in \cref{fig:lamg} also demonstrates the effect of the proposed CDC.
We specifically validate our assumption in \eqref{eq:cdc_rs}, which allows us to omit the global mean. The ``CDC w/ mean'' model includes the global mean, denoted as $\mbf{C}^s = \sm\left(\mbf{W}^{c_s}(\mbf{Q}^{c_s}\times\mbf{K}^{c_s\top} - \overline{\mbf{Q}^{c_s}\times\mbf{K}^{c_s\top}})\right)$.
We can note that the PSNR, SSIM, and DI are all getting lower with the global mean, and \cref{fig:lamf} demonstrates a smaller contribution area than \cref{fig:lamg}.
\begin{table}
\caption{Influence of Concerto Self-Attention. We analyze the model size and metrics from spatial Concertino $\mbf{C}^s$ to our method step by step.}
    \centering
\resizebox{0.96\linewidth}{!}{  
    \begin{tabular}{rccccc}
    \toprule
&\mc{3}{Metrics} &\mc{2}{Model Size}\\
&PSNR & SSIM &DI  &FLOPs(G) &Param(M)\\\midrule
Spatial Concertino $\mbf{C}^s$ &32.58  &0.953  &20.51 &\mr{4}{119.34}  &\mr{4}{21.2}\\
$\mbf{C}^s$ w/ Shifted Windows &32.84 &0.950 &24.61 &&\\
Prototype          &32.50  &0.952  &27.08 &  &\\
Concerto           &33.16 &0.958 &36.90 &  & \\\cmidrule{5-6}
CDC w/ mean        &33.23 &0.959 &34.62 &\mr{2}{119.77} &\mr{2}{22.5}\\
CDC                &\fst{33.30}  &\fst{0.960} &\fst{39.15} &  & \\\bottomrule
    \end{tabular}}
    \label{tab:concerto}
    \vspace{-3mm}
\end{table}
\begin{table}
    \centering
\caption{Analysis of Concerto Self-Attention. The \textit{Restormer-CSA} model uses the CSA instead of transposed self-attention.}
\resizebox{0.96\linewidth}{!}{
    \begin{tabular}{rcccccc}
    \toprule
\mr{2}{Method} &\mc{2}{\dataset{GoPro}~\cite{nah2017deep}} &\mc{2}{\dataset{HIDE}~\cite{shen2019human}} &\mc{2}{Model Size}\\
&PSNR &SSIM &PSNR &SSIM &FLOPs(G) &Param(M)\\ \midrule
Restormer \cite{zamir2022restormer} 	 &32.92	&0.961	&31.22	&0.942	&140.99 &26.1\\ 
\textit{Restormer-CSA}                      &\fst{33.32} &\fst{0.964}  &\fst{31.42}  &\fst{0.945}  &140.29 &31.0\\
\bottomrule
    \end{tabular}}
    \label{tab:restormer}
\end{table}

To evaluate the robustness and performance of our approach, we adopt the representative Restormer as the backbone and replace its transposed self-attention with our proposed CSA, denoted as \textit{Restormer-CSA} in \Cref{tab:restormer}. Following the same training procedures and model configurations as the original Restormer, \textit{Restormer-CSA} achieves a PSNR improvement of $0.4$ dB on \textsc{GoPro} and $0.2$ dB on \textsc{HIDE}, while reducing FLOPs by $0.5\%$.

\vspace{-10pt}
\section{Conclusion}
\vspace{-5pt}
We introduce an efficient {\ours} that addresses a primary limitation in previous vision Transformers (ViTs): the trade-off between global contextual representation and computational complexity. The proposed Concerto Self-Attention mechanism captures both local and global connectivity across spatial and channel domains with linear complexity. Additionally, we propose a gated-dconv MLP, which consolidates the traditional two-stage Transformer architecture into a single stage. Experimental results demonstrate that the gdMLP with CSA performs favorably compared to the two-staged design at lower FLOPs. By replacing the transposed self-attention with CSA, our method achieves higher PSNR at lower FLOPs compared to the canonical Restormer. Beyond single-image motion deblurring, {\ours} also performs favorably in deblurring with JPEG artifacts and deraining.

\section*{Acknowledgments}
This research was supported in part by the Institute of Information \& Communications Technology Planning \& Evaluation (IITP) grant funded by the Korean Government (MSIT) (No. RS-2024-00457882, National AI Research Lab Project).

{
    \small
    \bibliographystyle{ieeenat_fullname}
    \bibliography{main}
}
\clearpage
\setcounter{page}{1}
\maketitlesupplementary

In this supplemental material, we first provide real noisy image denoising results and discuss limitations. Then we report training details of tasks other than single-image motion deblurring.
Finally, we provide more visual comparisons.

\section{Evaluations on Real Noisy Image Denoising}
We further evaluate our method on the real noisy image denoising on the \dataset{SIDD} dataset. \Cref{tab:realdn} shows that the proposed methods, i.e., {\ours} and {\ours\textsuperscript{\dag}}\footnote{The architecture of {\ours} is detailed in \Cref{sec:detail}, while for {\ours\textsuperscript{\dag}}, the configuration of $L_1 - L_7$ is $[2, 4, 8, 16, 8, 4, 4]$.}, achieve comparable performance against state-of-the-art ones. 

\section{Limitation Analysis}
We have demonstrated the efficiency of {\ours} in the main paper. Although we propose a building block that can be applied to existing restoration models to solve various image restoration tasks, the backbone restoration model still requires a careful design for better performance improvement when using the proposed {\ours}. For example, the improvement of the proposed method on image denoising is marginal as shown in \Cref{tab:realdn}. 
\begin{table*}[hbt!]
    \centering
    \caption{Real noisy image denoising. * denotes methods using additional training data. \ours\textsuperscript{\dag} has more blocks in the latent layer.}
\resizebox{\textwidth}{!}{
    \begin{tabular}{cccccccccccccccccccc}\toprule
\mr{2}{Dataset}&\mr{2}{Metrics} &DnCNN &BM3D &CBDNet* &RIDNet* &AINDNet* &VDN &SADNet* &DANet+* &CycleISP* &MIRNet &DeamNet* &MPRNet &DAGL &Uformer &Restormer &NAFNet &\mr{2}{\ours} &\mr{2}{\ours\textsuperscript{\dag}} \\
               &              &\cite{zhang2017beyond} 	 &\cite{dabov2007image}   &\cite{guo2019toward}  &\cite{anwar2019real} &\cite{kim2020transfer}&\cite{yue2019variational} &\cite{chang2020spatial}   &\cite{yue2020dual}         &\cite{zamir2020cycleisp}  &\cite{zamir2020learning}  &\cite{ren2021adaptive}        &\cite{zamir2021multi} &\cite{mou2021dynamic}&\cite{wang2022uformer}    &\cite{zamir2022restormer}&\cite{chen2022simple}  & & \\\midrule
SIDD                      &PSNR &23.66 &25.65 &30.78 &38.71 &38.95  &39.28 &39.46 &39.47 &39.52 &39.72 &39.35 &39.71 &38.94 &39.77 &40.02  &\snd{40.30} &40.28 &\fst{40.33} \\
\cite{abdelhamed2018high} &SSIM & 0.583& 0.685& 0.801& 0.951& 0.952 &0.956 &0.957 &0.957 &0.957 &0.959 &0.955 &0.958 &0.953 &0.959 &0.960  &\snd{0.962} &0.962 &\fst{0.962} \\ \bottomrule
    \end{tabular}}
        \label{tab:realdn}
\end{table*}
\section{Evaluations on Single-Image Super-Resolution}
Although we do not design a dedicated model for super-resolution, our method yields competitive results compared to the state-of-the-art works. We replace the W-MSA and SW-MSA modules in the SwinIR with Concerto SA (Fig.~\ref{fig:model}) for Table~\ref{tab:sr}.
\begin{table}[hbt!]
    \centering
    \caption{Single-image super-resolution.}
\resizebox{\textwidth}{!}{
    \begin{tabular}{rccccccccccccc}
    \toprule
\mr{2}{Method} &\mr{2}{scale} &\mc{2}{\dataset{Set5}} &\mc{2}{\dataset{Set14}} &\mc{2}{\dataset{B100}} &\mc{2}{\dataset{Urban100}} &\mc{2}{\dataset{Manga109}}&\mc{2}{Avg.}\\
&&PSNR &SSIM &PSNR &SSIM &PSNR &SSIM&PSNR &SSIM &PSNR &SSIM&PSNR &SSIM\\ \midrule
EDSR \cite{lim2017enhanced} &\mr{5}{$\times 2$}  &38.11 &0.960 &33.92 &0.920 &32.32 &0.901 &32.93 &0.935 &39.10 &0.977 &35.28&0.939\\
NLSN \cite{mei2021image} &  &38.34 &0.962 &34.08 &0.923 &32.43 &0.903 &33.42 &0.939 &39.59 &0.979&35.57&0.941\\
ENLCN \cite{xia2022efficient}& &\fst{38.37} &\fst{0.962} &34.17 &0.923 &\fst{32.49} &\fst{0.903} &\fst{33.56}&\fst{0.940} &\snd{39.64} &\snd{0.979}&\fst{35.65} &\snd{0.941}\\
SwinIR \cite{liang2021swinir} & &\snd{38.35} &\snd{0.962} &\snd{34.14} &\snd{0.923} &32.44 &0.903 &33.40 &0.939 &39.60 &0.979&35.59 &0.941\\
CSA & &38.33 &0.962 &\fst{34.29} &\fst{0.924} &\snd{32.45} &\snd{0.903} &\snd{33.43} &\snd{0.940} &\fst{39.65} &\fst{0.979}&\snd{35.63} &\fst{0.942}\\\midrule
EDSR \cite{lim2017enhanced}&\mr{5}{$\times 4$}  & 32.46 &0.897 &28.80 &0.788 &27.71 & 0.742 &26.64 &0.803 &31.02 &0.915 &29.33&0.829\\
NLSN \cite{mei2021image} &  &32.59 &0.900 &28.87 &0.789 &27.78 &0.744 &26.96 &0.811 &31.27 &0.918&29.49&0.832\\
ENLCN \cite{xia2022efficient}& &32.67 &0.900 &28.94 &0.789 &27.82 &0.745 &\fst{27.12} &\fst{0.814} &31.33 &0.919&29.58&0.833\\
SwinIR \cite{liang2021swinir}&&\snd{32.72} &\snd{0.902} &\snd{28.94} &\snd{0.791} &\fst{27.83} &\fst{0.746} &\snd{27.07} &\snd{0.816} &\fst{31.67} &\fst{0.923} &\fst{29.65}&\fst{0.836}\\
CSA & &\fst{32.74} &\fst{0.902} &\fst{28.96} &\fst{0.791} &\snd{27.82} &\snd{0.746} &26.92 &0.811 &\snd{31.49} &\snd{0.922}&\snd{29.59}&\snd{0.834}\\
\bottomrule
\end{tabular}}
\label{tab:sr}
\end{table}
\section{Running Time Analysis}
In this section, we compare the deblurring methods of self-attention on HIDE \cite{shen2019human}. As shown in Table~\ref{tab:time}, our methods generate the highest scores with relatively short running times. This table demonstrates the efficiency of our Concerto self-attention. The running times are calculated on one RTX 4080 Super GPU with one $256\times256\times3$ random input image.

\begin{table}
    \centering
    \caption{Running time analysis. We compare our model to self-attention-based deblurring methods. The HIDE \cite{shen2019human} dataset is used for metrics. * denotes the model without Test-time Local Converter.}
    \resizebox{\textwidth}{!}{
    \begin{tabular}{ccccccccc}
    \toprule
                & Uformer \cite{wang2022uformer} & Stripformer\cite{tsai2022stripformer} & Restormer\cite{zamir2022restormer} & Restormer-local\cite{chu2022improving} & GRL\cite{li2023efficient}   & FFTformer\cite{kong2023efficient} & \ourslite* & \ours* \\\midrule 
        PSNR    & 30.90   & 31.03       & 31.22     & 31.49           & 31.65 & 31.62     & \snd{31.67}      & \fst{31.96}  \\
        SSIM    & 0.953   & 0.940       & 0.942     & 0.945           & 0.947 & 0.946     & \snd{0.947}      & \fst{0.950}  \\
        FLOPs (G)  & \fst{90.16}   & 177.43      &140        &155.13           &1,289  &139.09     &\snd{116.79}            & 220.2 \\
        Param (M)   & 50.8    & \snd{19.7}        &26.1       &26.1             &20.2   &\fst{16.6}       &28.9              &50.5 \\
        Time (ms) & \fst{22.47}   &\snd{25.67}        & 57.60     &183.28           &344.60 & 103.26    & 45.03      & 83.46  \\\bottomrule
    \end{tabular}}
    \label{tab:time}
\end{table}

\renewcommand{\thesubfigure}{\thefigure.\arabic{subfigure}}
\captionsetup[subfigure]{labelformat=empty}
\begin{figure}
    \centering
    \subfloat[]{\includegraphics[width=0.5\linewidth]{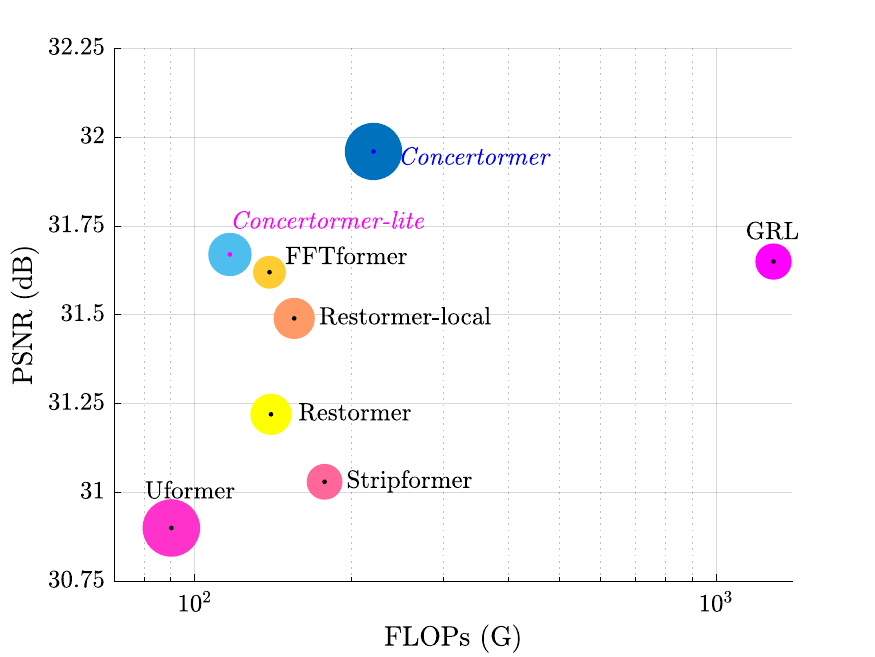}}
    \subfloat[]{\includegraphics[width=0.5\linewidth]{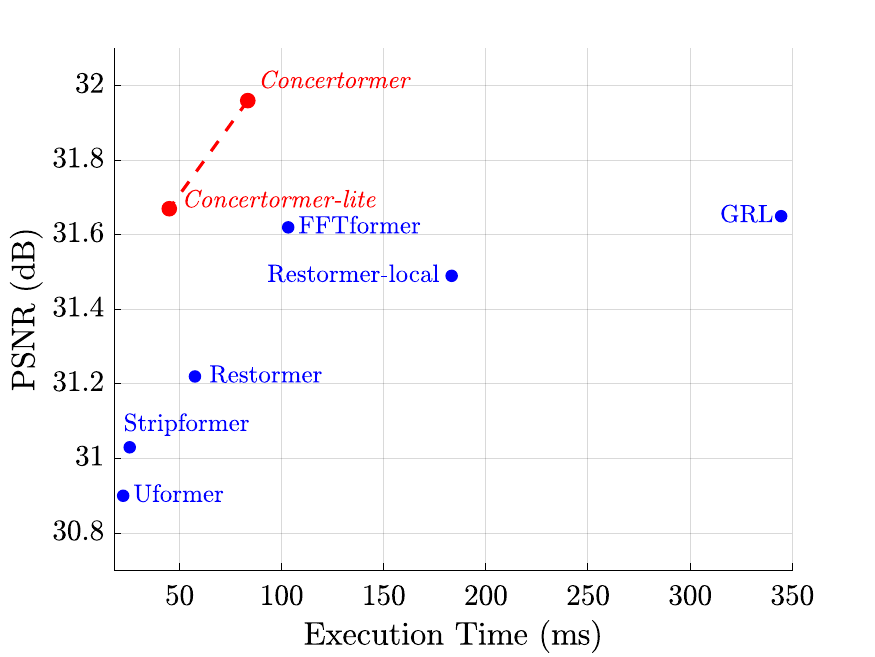}}
    \caption{Visualization of model size and execution time.}
    \label{fig:supp_size&time}
\end{figure}

\section{Other Training Details}
For real noisy image denoising on the \dataset{SIDD} dataset, we train the model for 400,000 iterations, following \cite{chen2022simple}, as additional iterations do not yield further improvements. Since both the training and testing data consist of $256 \times 256$ pixel images, we do not employ progressive training; instead, we train the model exclusively on $256 \times 256$ patches.

For the deraining task, however, we adopt a progressive training strategy. The model is trained with $192 \times 192$ patches for 100,000 iterations, followed by $256 \times 256$ patches for 200,000 iterations, $320 \times 320$ patches for 100,000 iterations, and an additional 10,000 iterations using $128 \times 128$ patches.

As for \dataset{REDS}, we train the model as described in \Cref{sec:data} with the configuration as \Cref{sec:detail}.
\section{More Visual Comparisons}
\label{sec:moreresults}
\renewcommand{\thesubfigure}{\alph{subfigure}}
\captionsetup[subfigure]{labelformat=parens}
\begin{figure*}[hbt!]
  \centering
  \tiny
  \begin{minipage}[b]{1.\linewidth}
\subfloat[Blurry Input]{
    \begin{minipage}{.195\linewidth} 
      \centering
      \includegraphics[width=\linewidth]{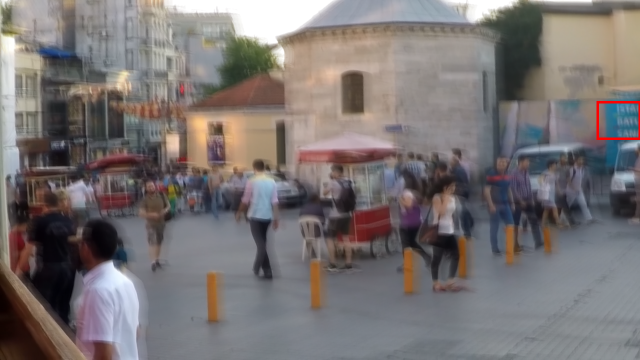}
      \includegraphics[width=\linewidth]{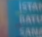}  	
    \end{minipage}
	}
\subfloat[MPRNet \cite{zamir2021multi}]{
    \begin{minipage}{.195\linewidth} 
      \centering
      \includegraphics[width=\linewidth]{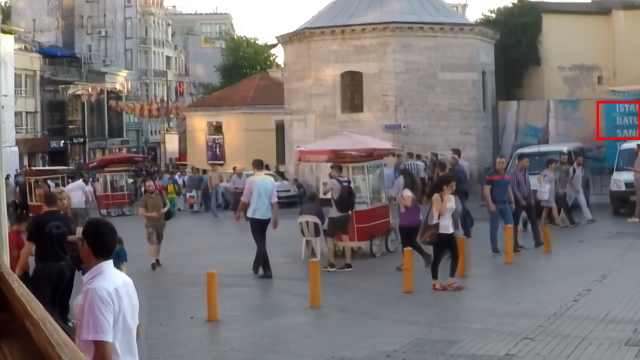}
      \includegraphics[width=\linewidth]{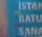}  	
    \end{minipage}
	}
\subfloat[Uformer \cite{wang2022uformer} ]{
    \begin{minipage}{.195\linewidth} 
      \centering
      \includegraphics[width=\linewidth]{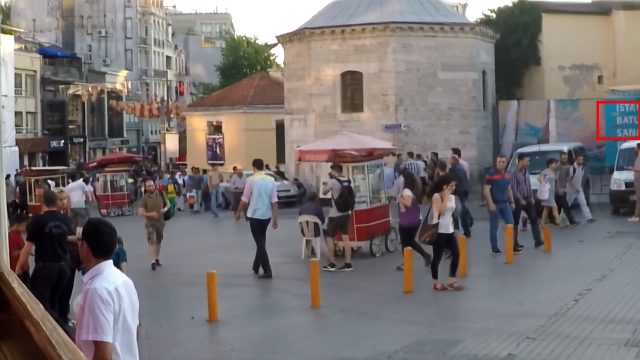}
      \includegraphics[width=\linewidth]{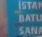}  	
    \end{minipage}
	}
\subfloat[Restormer \cite{zamir2022restormer} ]{
    \begin{minipage}{.195\linewidth} 
      \centering
      \includegraphics[width=\linewidth]{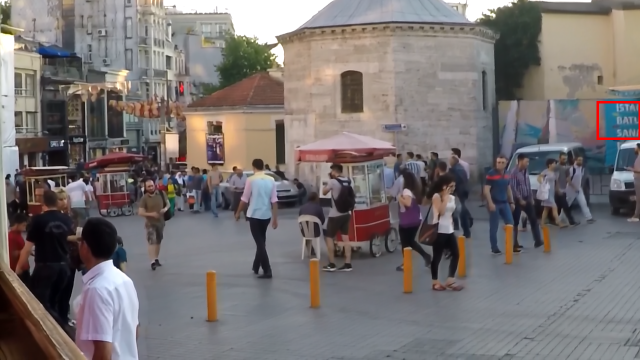}
      \includegraphics[width=\linewidth]{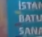}  	
    \end{minipage}
	}
\subfloat[Stripformer \cite{tsai2022stripformer} ]{
    \begin{minipage}{.195\linewidth} 
      \centering
      \includegraphics[width=\linewidth]{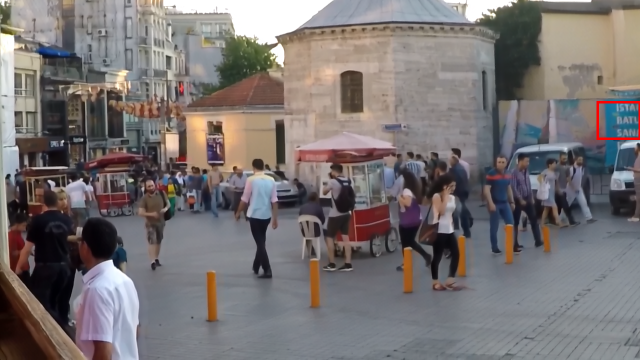}
      \includegraphics[width=\linewidth]{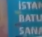}  	
    \end{minipage}
	}\\
\subfloat[GRL \cite{li2023efficient} ]{
    \begin{minipage}{.195\linewidth} 
      \centering
      \includegraphics[width=\linewidth]{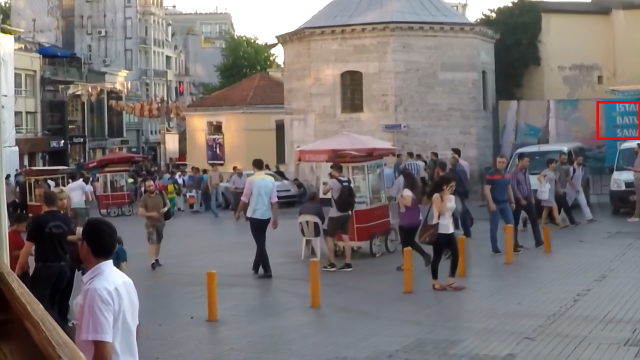}
      \includegraphics[width=\linewidth]{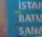}  	
    \end{minipage}
	} 
\subfloat[FFTformer \cite{kong2023efficient} ]{
    \begin{minipage}{.195\linewidth} 
      \centering
      \includegraphics[width=\linewidth]{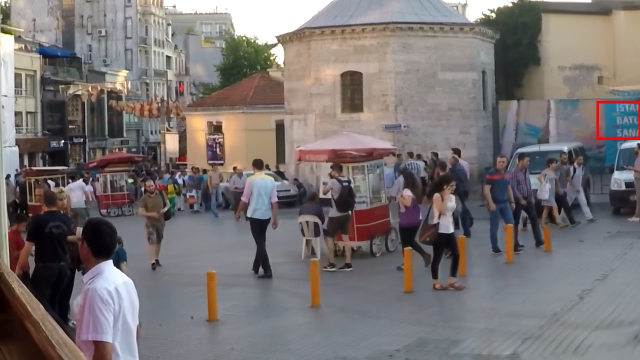}
      \includegraphics[width=\linewidth]{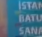}  	
    \end{minipage}
	}
\subfloat[DiffIR \cite{xia2023diffir}]{
    \begin{minipage}{.195\linewidth} 
      \centering
      \includegraphics[width=\linewidth]{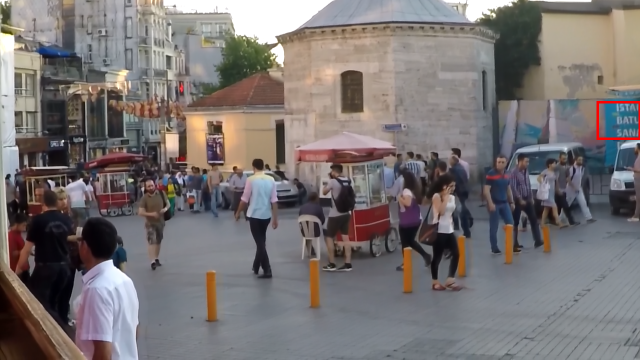}
      \includegraphics[width=\linewidth]{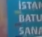}  	
    \end{minipage}
	}
\subfloat[{\ours}]{
    \begin{minipage}{.195\linewidth} 
      \centering
      \includegraphics[width=\linewidth]{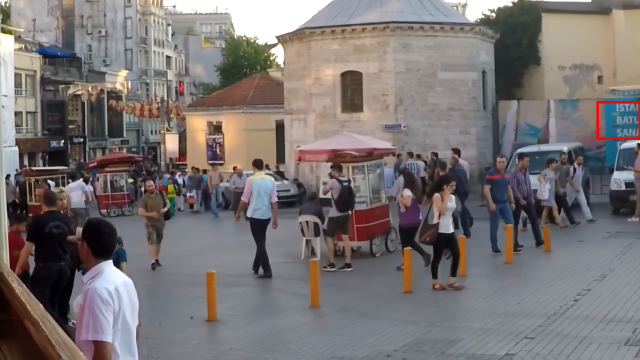}
      \includegraphics[width=\linewidth]{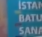}  	
    \end{minipage}
	}
\subfloat[GT]{
    \begin{minipage}{.195\linewidth} 
      \centering
      \includegraphics[width=\linewidth]{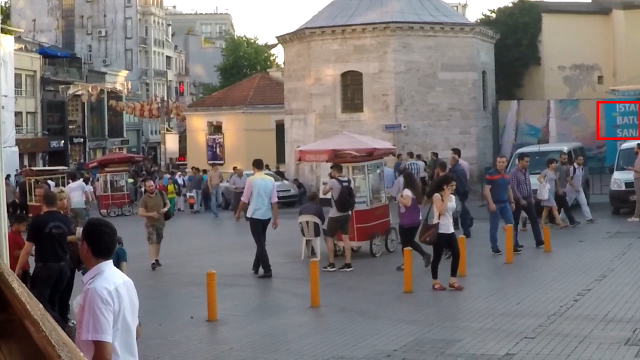}
      \includegraphics[width=\linewidth]{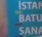}  	
    \end{minipage}
	}\hfill
  \end{minipage}
	\vfill
      \caption{Visual comparisons of single-image motion deblurring on \dataset{GoPro} \cite{nah2017deep}.}
\label{fig:s_gopro}
\end{figure*}
\begin{figure*}[hbt!]
  \centering
  \tiny
  \begin{minipage}[b]{1.\linewidth}
\subfloat[Blurry Input]{
    \begin{minipage}{.195\linewidth} 
      \centering
      \includegraphics[width=\linewidth]{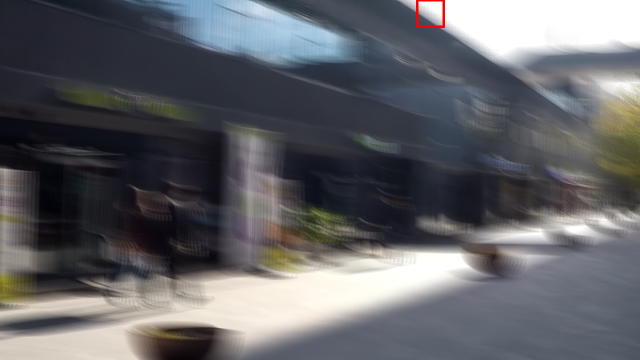}
      \includegraphics[width=\linewidth]{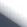}  	
    \end{minipage}
	}
\subfloat[MPRNet \cite{zamir2021multi}]{
    \begin{minipage}{.195\linewidth} 
      \centering
      \includegraphics[width=\linewidth]{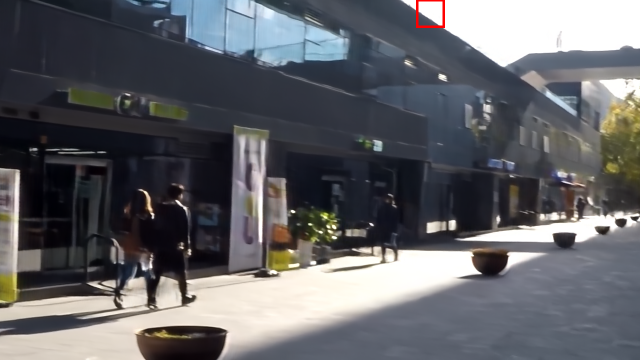}
      \includegraphics[width=\linewidth]{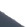}  	
    \end{minipage}
	}
\subfloat[Uformer \cite{wang2022uformer} ]{
    \begin{minipage}{.195\linewidth} 
      \centering
      \includegraphics[width=\linewidth]{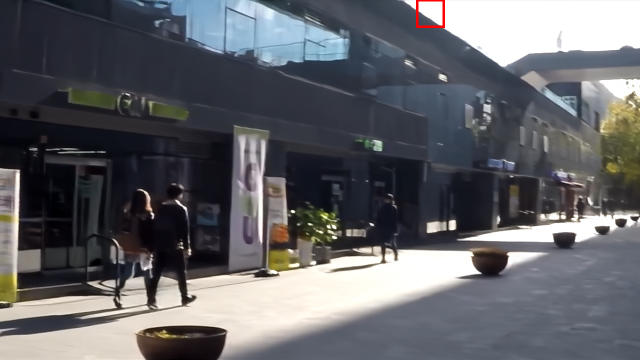}
      \includegraphics[width=\linewidth]{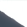}  	
    \end{minipage}
	}
\subfloat[Restormer \cite{zamir2022restormer} ]{
    \begin{minipage}{.195\linewidth} 
      \centering
      \includegraphics[width=\linewidth]{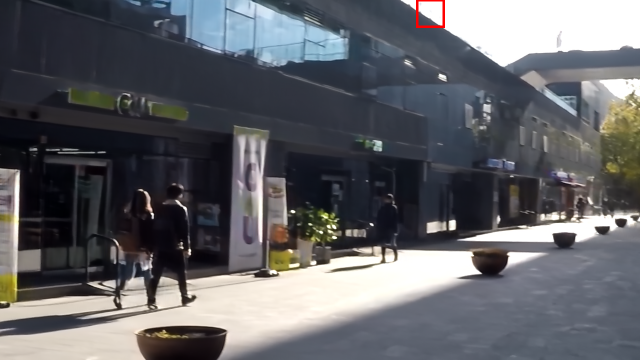}
      \includegraphics[width=\linewidth]{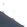}  	
    \end{minipage}
	}
\subfloat[Stripformer \cite{tsai2022stripformer} ]{
    \begin{minipage}{.195\linewidth} 
      \centering
      \includegraphics[width=\linewidth]{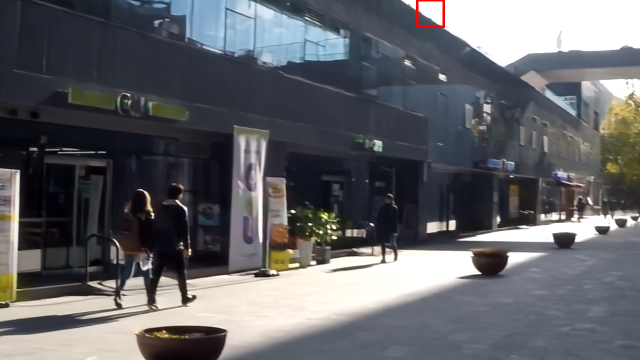}
      \includegraphics[width=\linewidth]{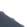}  	
    \end{minipage}
	}\\
\subfloat[GRL \cite{li2023efficient} ]{
    \begin{minipage}{.195\linewidth} 
      \centering
      \includegraphics[width=\linewidth]{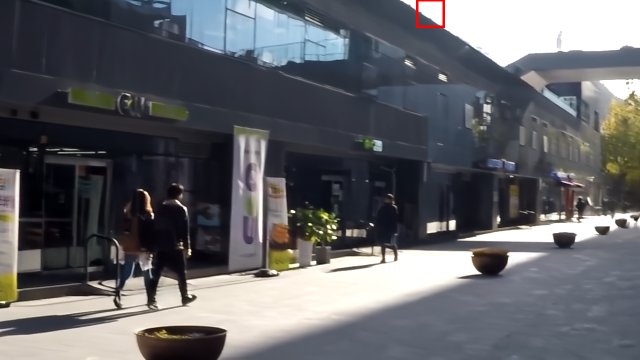}
      \includegraphics[width=\linewidth]{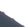}  	
    \end{minipage}
	} 
\subfloat[FFTformer \cite{kong2023efficient} ]{
    \begin{minipage}{.195\linewidth} 
      \centering
      \includegraphics[width=\linewidth]{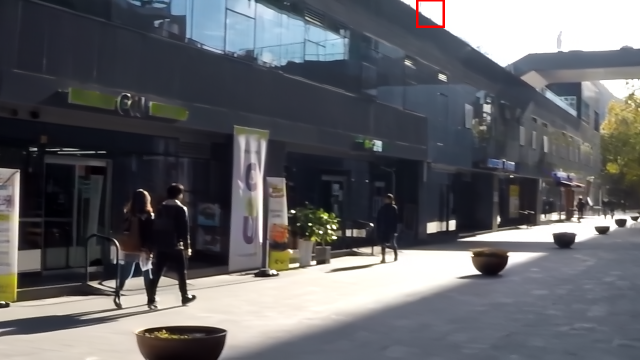}
      \includegraphics[width=\linewidth]{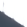}  	
    \end{minipage}
	}
\subfloat[DiffIR \cite{xia2023diffir}]{
    \begin{minipage}{.195\linewidth} 
      \centering
      \includegraphics[width=\linewidth]{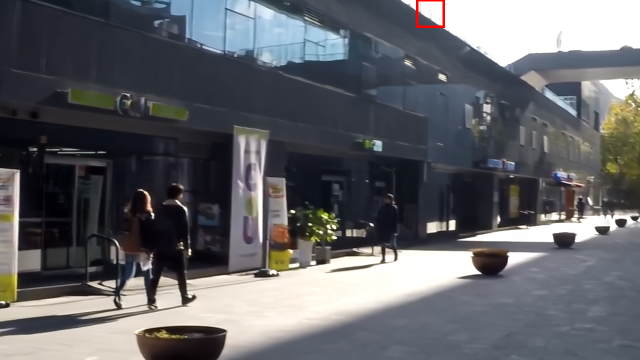}
      \includegraphics[width=\linewidth]{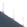}  	
    \end{minipage}
	}
\subfloat[{\ours}]{
    \begin{minipage}{.195\linewidth} 
      \centering
      \includegraphics[width=\linewidth]{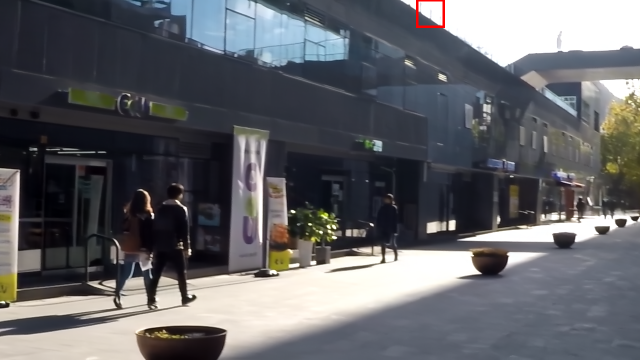}
      \includegraphics[width=\linewidth]{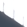}  	
    \end{minipage}
	}
\subfloat[GT]{
    \begin{minipage}{.195\linewidth} 
      \centering
      \includegraphics[width=\linewidth]{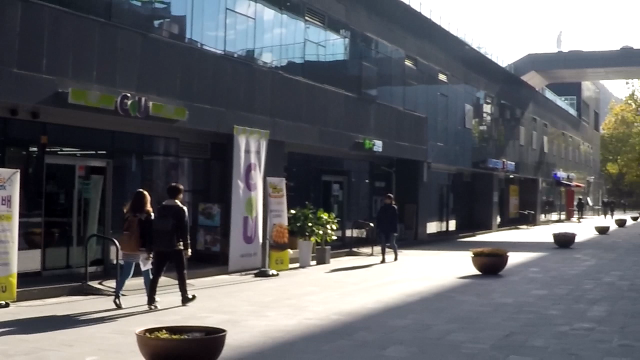}
      \includegraphics[width=\linewidth]{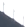}  	
    \end{minipage}
	}\hfill
  \end{minipage}
	\vfill
      \caption{Visual comparisons of single-image motion deblurring on \dataset{GoPro} \cite{nah2017deep}.}
\label{fig:s_gopro2}
\end{figure*}
\begin{figure*}[hbt!]
  \centering
  \tiny
  \begin{minipage}[b]{1.\linewidth}
\subfloat[Blurry Input]{
    \begin{minipage}{.195\linewidth} 
      \centering
      \includegraphics[width=\linewidth]{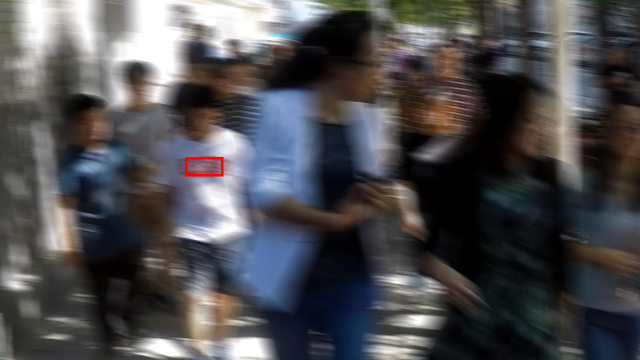}
      \includegraphics[width=\linewidth]{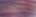}  	
    \end{minipage}
	}
\subfloat[MPRNet \cite{zamir2021multi}]{
    \begin{minipage}{.195\linewidth} 
      \centering
      \includegraphics[width=\linewidth]{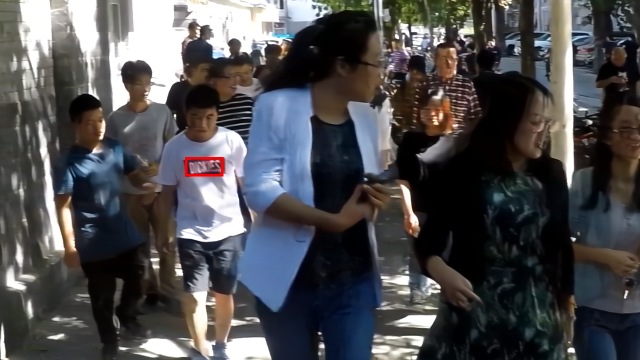}
      \includegraphics[width=\linewidth]{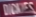}  	
    \end{minipage}
	}
\subfloat[Uformer \cite{wang2022uformer} ]{
    \begin{minipage}{.195\linewidth} 
      \centering
      \includegraphics[width=\linewidth]{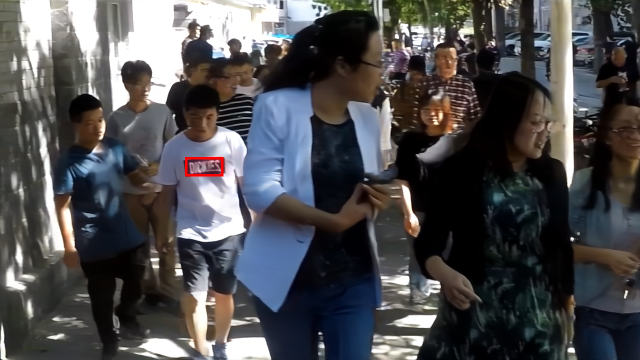}
      \includegraphics[width=\linewidth]{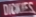}  	
    \end{minipage}
	}
\subfloat[Restormer \cite{zamir2022restormer} ]{
    \begin{minipage}{.195\linewidth} 
      \centering
      \includegraphics[width=\linewidth]{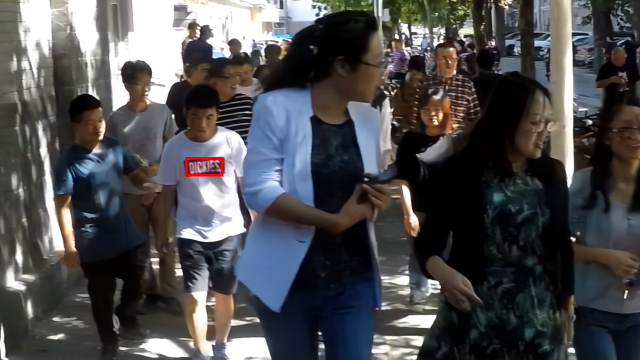}
      \includegraphics[width=\linewidth]{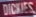}  	
    \end{minipage}
	}
\subfloat[Stripformer \cite{tsai2022stripformer} ]{
    \begin{minipage}{.195\linewidth} 
      \centering
      \includegraphics[width=\linewidth]{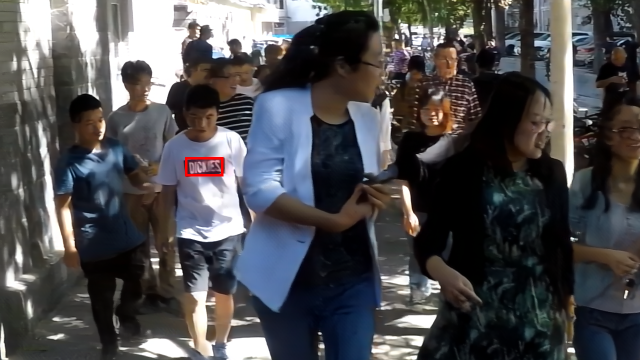}
      \includegraphics[width=\linewidth]{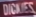}  	
    \end{minipage}
	}\\
\subfloat[GRL \cite{li2023efficient} \label{fig:hidegrl}]{
    \begin{minipage}{.195\linewidth} 
      \centering
      \includegraphics[width=\linewidth]{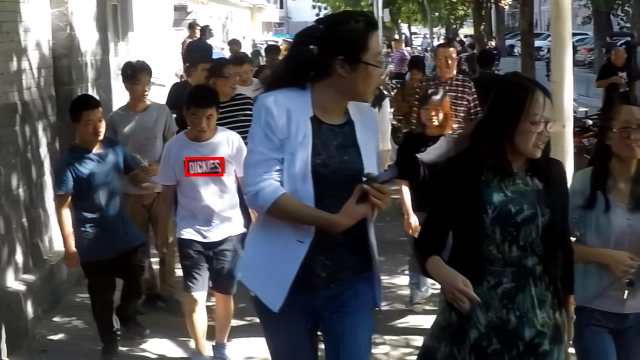}
      \includegraphics[width=\linewidth]{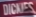}  	
    \end{minipage}
	}
\subfloat[FFTformer \cite{kong2023efficient} ]{
    \begin{minipage}{.195\linewidth} 
      \centering
      \includegraphics[width=\linewidth]{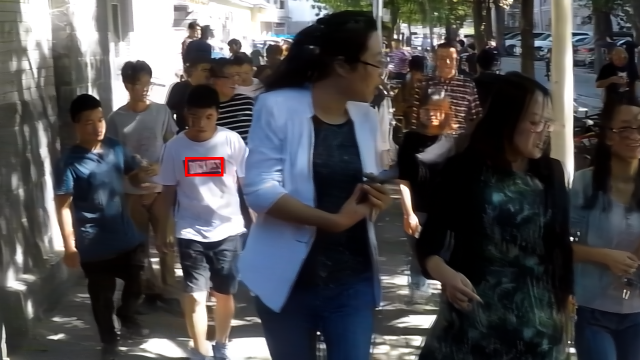}
      \includegraphics[width=\linewidth]{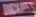}  	
    \end{minipage}
	}
\subfloat[DiffIR \cite{xia2023diffir}]{
    \begin{minipage}{.195\linewidth} 
      \centering
      \includegraphics[width=\linewidth]{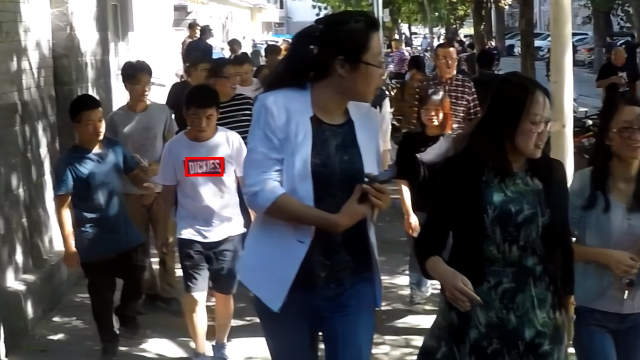}
      \includegraphics[width=\linewidth]{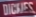}  	
    \end{minipage}
	}
\subfloat[\ours]{
    \begin{minipage}{.195\linewidth} 
      \centering
      \includegraphics[width=\linewidth]{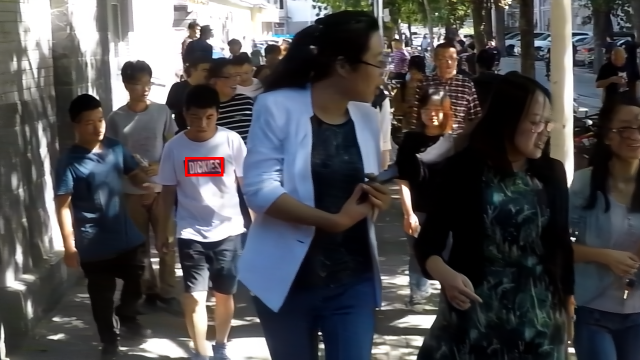}
      \includegraphics[width=\linewidth]{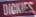}  	
    \end{minipage}
	}
\subfloat[GT]{
    \begin{minipage}{.195\linewidth} 
      \centering
      \includegraphics[width=\linewidth]{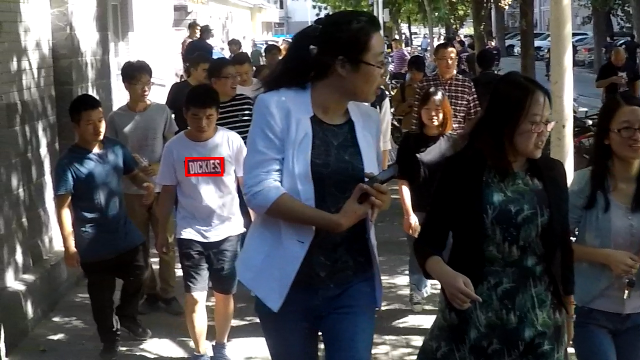}
      \includegraphics[width=\linewidth]{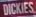}  	
    \end{minipage}
	}\hfill
  \end{minipage}
	\vfill
      \caption{Visual comparisons of single-image motion deblurring on \dataset{HIDE} \cite{shen2019human}.}
\label{fig:s_hide}
\end{figure*}
\begin{figure*}[hbt!]
  \centering
  \tiny
  \begin{minipage}[b]{1.\linewidth}
\subfloat[Blurry Input]{
    \begin{minipage}{.195\linewidth} 
      \centering
      \includegraphics[width=\linewidth]{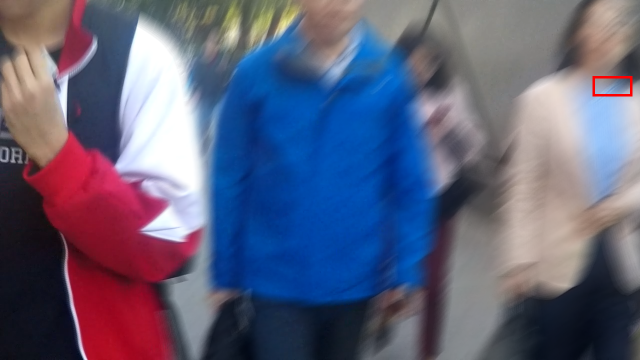}
      \includegraphics[width=\linewidth]{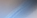}  	
    \end{minipage}
	}
\subfloat[MPRNet \cite{zamir2021multi}]{
    \begin{minipage}{.195\linewidth} 
      \centering
      \includegraphics[width=\linewidth]{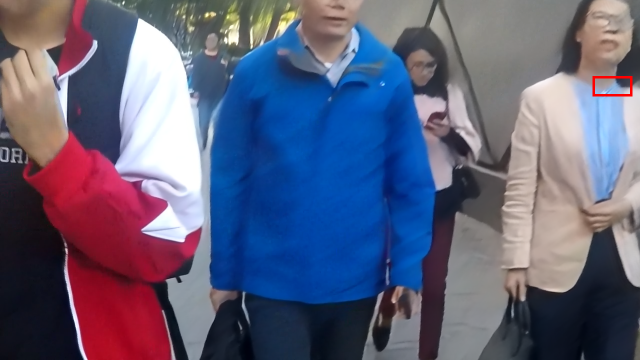}
      \includegraphics[width=\linewidth]{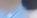}  	
    \end{minipage}
	}
\subfloat[Uformer \cite{wang2022uformer} ]{
    \begin{minipage}{.195\linewidth} 
      \centering
      \includegraphics[width=\linewidth]{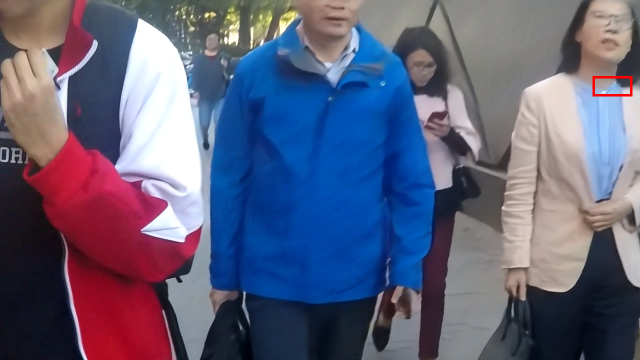}
      \includegraphics[width=\linewidth]{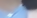}  	
    \end{minipage}
	}
\subfloat[Restormer \cite{zamir2022restormer} ]{
    \begin{minipage}{.195\linewidth} 
      \centering
      \includegraphics[width=\linewidth]{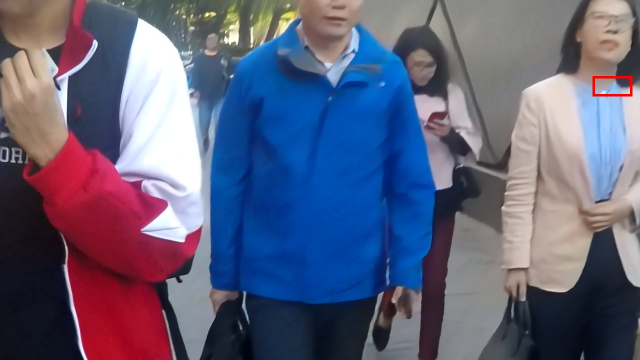}
      \includegraphics[width=\linewidth]{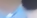}  	
    \end{minipage}
	}
\subfloat[Stripformer \cite{tsai2022stripformer} ]{
    \begin{minipage}{.195\linewidth} 
      \centering
      \includegraphics[width=\linewidth]{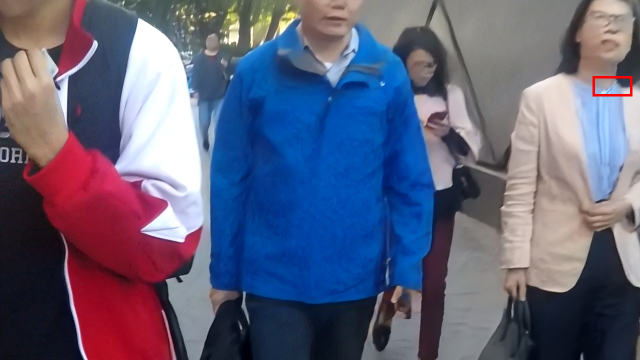}
      \includegraphics[width=\linewidth]{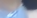}  	
    \end{minipage}
	}\\
\subfloat[GRL \cite{li2023efficient} \label{fig:hide2grl}]{
    \begin{minipage}{.195\linewidth} 
      \centering
      \includegraphics[width=\linewidth]{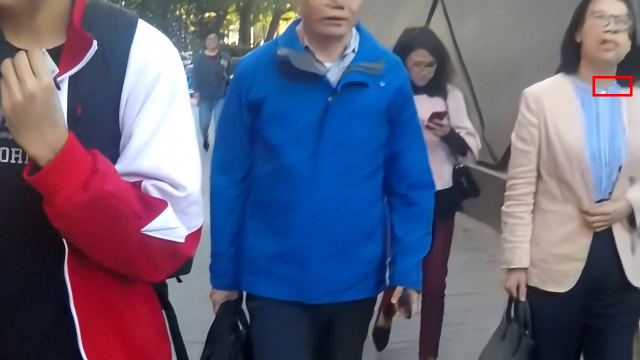}
      \includegraphics[width=\linewidth]{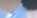}  	
    \end{minipage}
	}
\subfloat[FFTformer \cite{kong2023efficient} ]{
    \begin{minipage}{.195\linewidth} 
      \centering
      \includegraphics[width=\linewidth]{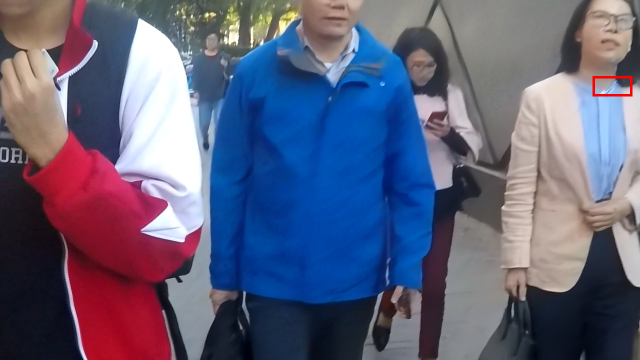}
      \includegraphics[width=\linewidth]{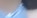}  	
    \end{minipage}
	}
\subfloat[DiffIR \cite{xia2023diffir}]{
    \begin{minipage}{.195\linewidth} 
      \centering
      \includegraphics[width=\linewidth]{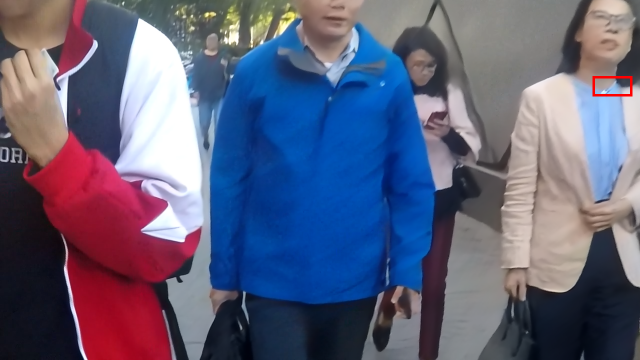}
      \includegraphics[width=\linewidth]{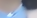}  	
    \end{minipage}
	}
\subfloat[\ours]{
    \begin{minipage}{.195\linewidth} 
      \centering
      \includegraphics[width=\linewidth]{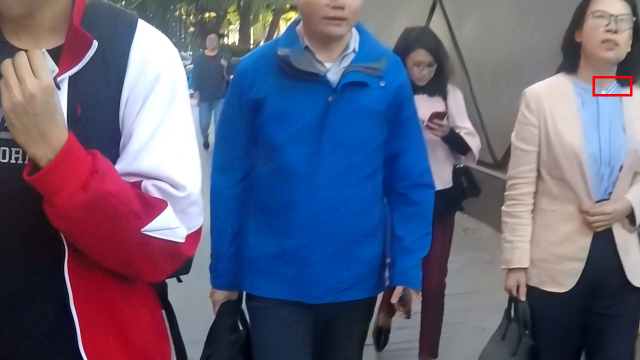}
      \includegraphics[width=\linewidth]{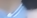}  	
    \end{minipage}
	}
\subfloat[GT]{
    \begin{minipage}{.195\linewidth} 
      \centering
      \includegraphics[width=\linewidth]{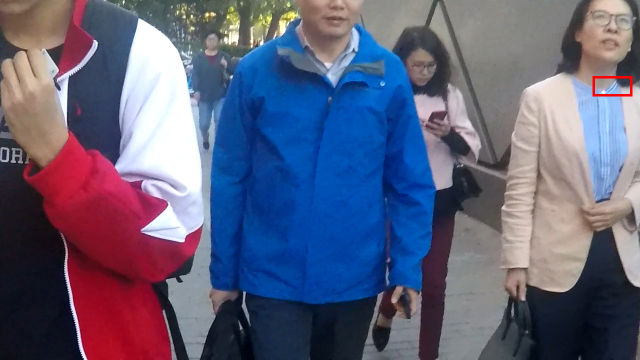}
      \includegraphics[width=\linewidth]{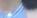}  	
    \end{minipage}
	}\hfill
  \end{minipage}
	\vfill
      \caption{Visual comparisons of single-image motion deblurring on \dataset{HIDE} \cite{shen2019human}.}
\label{fig:s_hide2}
\end{figure*}
\begin{figure*}[hbt!]
\vspace{-30pt}
  \centering
  \tiny
  \begin{minipage}[b]{1.\linewidth}
\subfloat[Blurry Input]{
    \begin{minipage}{.195\linewidth} 
      \centering
      \includegraphics[width=\linewidth]{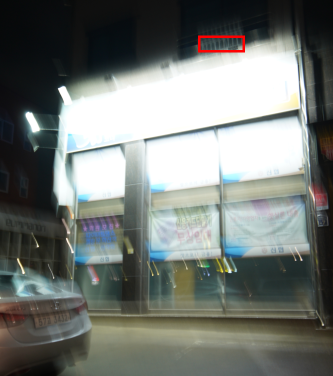}
      \includegraphics[width=\linewidth]{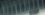}  	
    \end{minipage}
	}
\subfloat[DeblurGAN-v2 \cite{kupyn2019deblurgan}]{
    \begin{minipage}{.195\linewidth} 
      \centering
      \includegraphics[width=\linewidth]{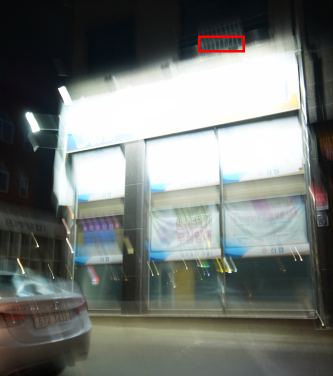}
      \includegraphics[width=\linewidth]{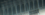}  	
    \end{minipage}
	}
\subfloat[MAXIM-3S \cite{tu2022maxim}]{
    \begin{minipage}{.195\linewidth} 
      \centering
      \includegraphics[width=\linewidth]{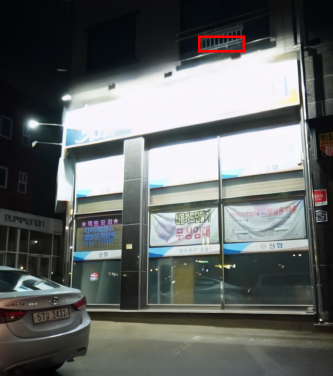}
      \includegraphics[width=\linewidth]{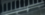}  	
    \end{minipage}
	}
\subfloat[DeepRFT+ \cite{mao2021deep}]{
    \begin{minipage}{.195\linewidth} 
      \centering
      \includegraphics[width=\linewidth]{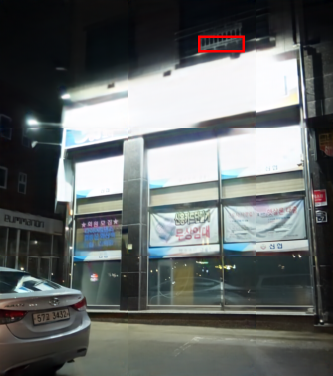}
      \includegraphics[width=\linewidth]{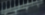}  	
    \end{minipage}
	}
\subfloat[Stripformer \cite{tsai2022stripformer} ]{
    \begin{minipage}{.195\linewidth} 
      \centering
      \includegraphics[width=\linewidth]{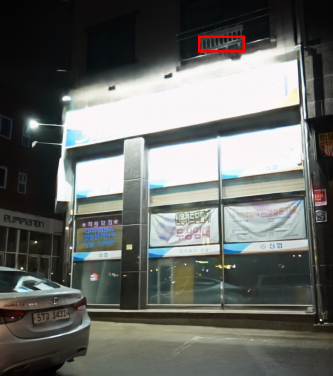}
      \includegraphics[width=\linewidth]{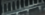}  	
    \end{minipage}
	}\\
\subfloat[FMIMO-UNet+ \cite{mao2023intriguing}]{
    \begin{minipage}{.195\linewidth} 
      \centering
      \includegraphics[width=\linewidth]{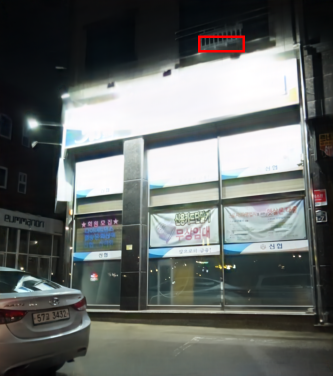}
      \includegraphics[width=\linewidth]{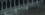}  	
    \end{minipage}
	}
\subfloat[FFTformer \cite{kong2023efficient} ]{
    \begin{minipage}{.195\linewidth} 
      \centering
      \includegraphics[width=\linewidth]{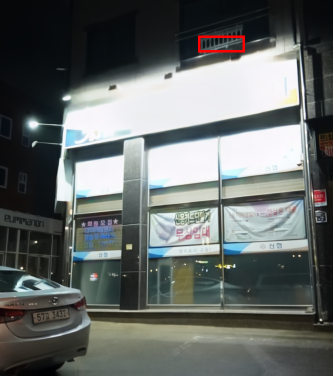}
      \includegraphics[width=\linewidth]{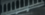}  	
    \end{minipage}
	}
\subfloat[GRL \cite{li2023efficient}]{
    \begin{minipage}{.195\linewidth} 
      \centering
      \includegraphics[width=\linewidth]{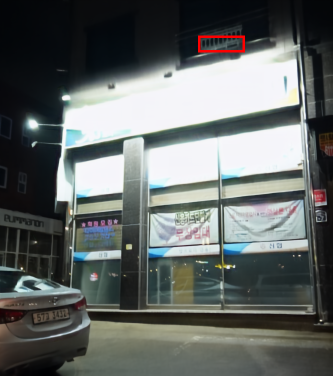}
      \includegraphics[width=\linewidth]{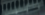}  	
    \end{minipage}
	}
\subfloat[\ours]{
    \begin{minipage}{.195\linewidth} 
      \centering
      \includegraphics[width=\linewidth]{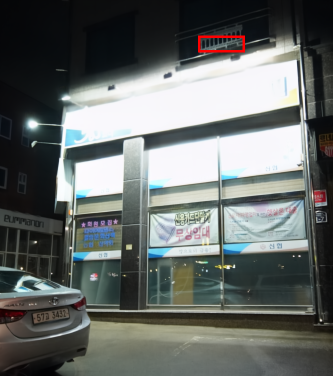}
      \includegraphics[width=\linewidth]{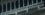}  	
    \end{minipage}
	}
\subfloat[GT]{
    \begin{minipage}{.195\linewidth} 
      \centering
      \includegraphics[width=\linewidth]{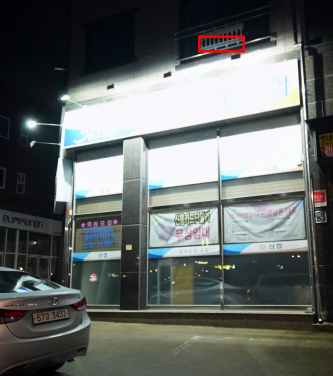}
      \includegraphics[width=\linewidth]{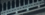}  	
    \end{minipage}
	}\hfill
  \end{minipage}
	\vfill
    \vspace{-3pt}
      \caption{Visual comparisons of single-image motion deblurring on \dataset{RealBlur\_J} \cite{rim2020real}.}
\label{fig:s_realj}
\end{figure*}
\begin{figure*}[hbt!]
\vspace{-20pt}
  \centering
  \tiny
  \begin{minipage}[b]{1.\linewidth}
\subfloat[Blurry Input]{
    \begin{minipage}{.195\linewidth} 
      \centering
      \includegraphics[width=\linewidth]{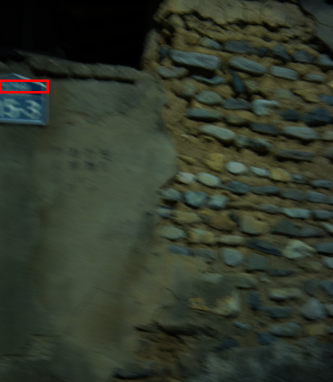}
      \includegraphics[width=\linewidth]{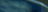}  	
    \end{minipage}
	}
\subfloat[DeblurGAN-v2 \cite{kupyn2019deblurgan}]{
    \begin{minipage}{.195\linewidth} 
      \centering
      \includegraphics[width=\linewidth]{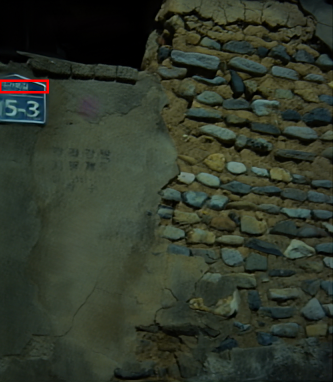}
      \includegraphics[width=\linewidth]{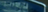}  	
    \end{minipage}
	}
\subfloat[MAXIM-3S \cite{tu2022maxim}]{
    \begin{minipage}{.195\linewidth} 
      \centering
      \includegraphics[width=\linewidth]{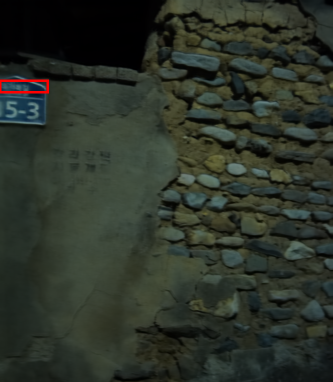}
      \includegraphics[width=\linewidth]{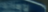}  	
    \end{minipage}
	}
\subfloat[DeepRFT+ \cite{mao2021deep}]{
    \begin{minipage}{.195\linewidth} 
      \centering
      \includegraphics[width=\linewidth]{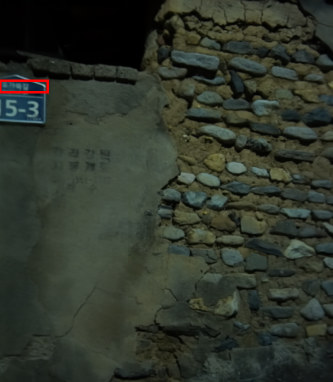}
      \includegraphics[width=\linewidth]{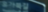}  	
    \end{minipage}
	}
\subfloat[Stripformer \cite{tsai2022stripformer} ]{
    \begin{minipage}{.195\linewidth} 
      \centering
      \includegraphics[width=\linewidth]{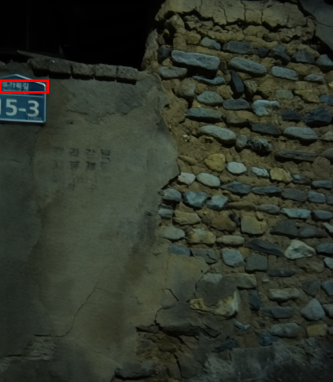}
      \includegraphics[width=\linewidth]{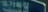}  	
    \end{minipage}
	}\\
\subfloat[FMIMO-UNet+ \cite{mao2023intriguing}]{
    \begin{minipage}{.195\linewidth} 
      \centering
      \includegraphics[width=\linewidth]{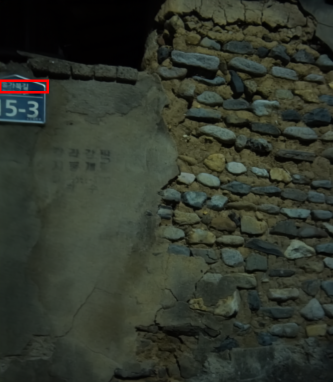}
      \includegraphics[width=\linewidth]{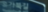}  	
    \end{minipage}
	}
\subfloat[FFTformer \cite{kong2023efficient} ]{
    \begin{minipage}{.195\linewidth} 
      \centering
      \includegraphics[width=\linewidth]{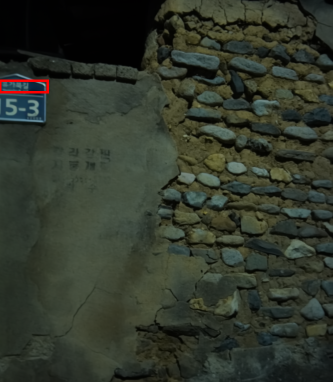}
      \includegraphics[width=\linewidth]{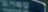}  	
    \end{minipage}
	}
\subfloat[GRL \cite{li2023efficient}]{
    \begin{minipage}{.195\linewidth} 
      \centering
      \includegraphics[width=\linewidth]{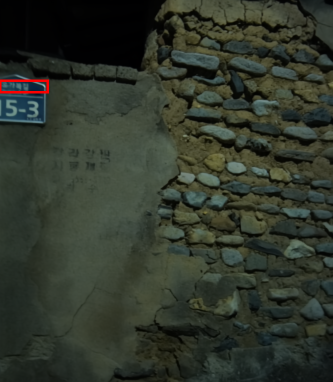}
      \includegraphics[width=\linewidth]{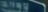}  	
    \end{minipage}
	}
\subfloat[\ours]{
    \begin{minipage}{.195\linewidth} 
      \centering
      \includegraphics[width=\linewidth]{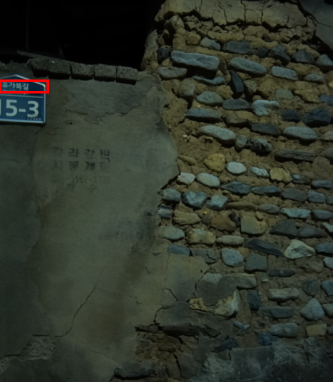}
      \includegraphics[width=\linewidth]{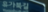}  	
    \end{minipage}
	}
\subfloat[GT]{
    \begin{minipage}{.195\linewidth} 
      \centering
      \includegraphics[width=\linewidth]{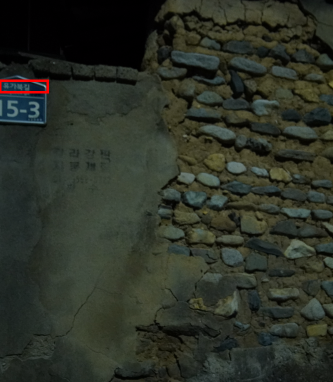}
      \includegraphics[width=\linewidth]{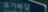}  	
    \end{minipage}
	}\hfill
  \end{minipage}
	\vfill
    \vspace{-3pt}
      \caption{Visual comparisons of single-image motion deblurring on \dataset{RealBlur\_J} \cite{rim2020real}.}
\label{fig:s_realj2}
\end{figure*}
\begin{figure*}[hbt!]
\vspace{-20pt}
  \centering
  \tiny
  \begin{minipage}[b]{1.\linewidth}
\subfloat[Blurry Input]{
    \begin{minipage}{.195\linewidth} 
      \centering
      \includegraphics[width=\linewidth]{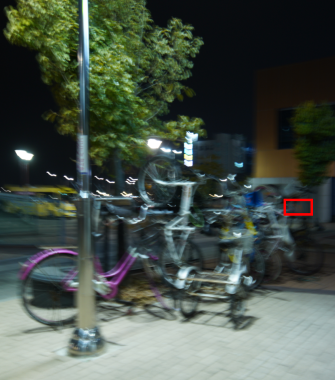}
      \includegraphics[width=\linewidth]{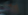}  	
    \end{minipage}
	}
\subfloat[DeblurGAN-v2 \cite{kupyn2019deblurgan}]{
    \begin{minipage}{.195\linewidth} 
      \centering
      \includegraphics[width=\linewidth]{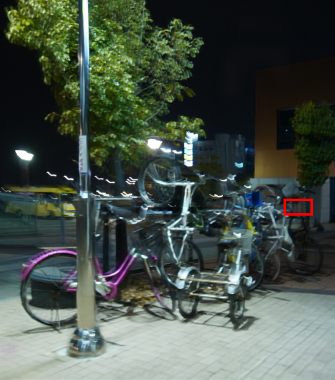}
      \includegraphics[width=\linewidth]{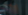}  	
    \end{minipage}
	}
\subfloat[MAXIM-3S \cite{tu2022maxim}]{
    \begin{minipage}{.195\linewidth} 
      \centering
      \includegraphics[width=\linewidth]{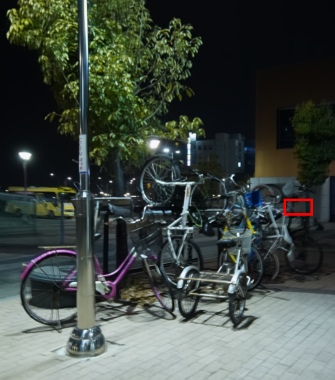}
      \includegraphics[width=\linewidth]{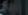}  	
    \end{minipage}
	}
\subfloat[DeepRFT+ \cite{mao2021deep}]{
    \begin{minipage}{.195\linewidth} 
      \centering
      \includegraphics[width=\linewidth]{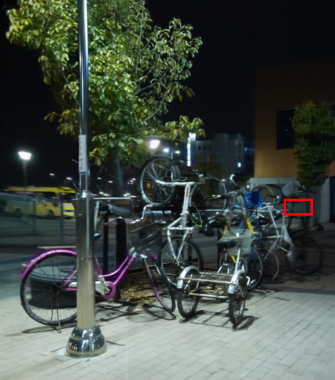}
      \includegraphics[width=\linewidth]{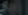}  	
    \end{minipage}
	}
\subfloat[Stripformer \cite{tsai2022stripformer} ]{
    \begin{minipage}{.195\linewidth} 
      \centering
      \includegraphics[width=\linewidth]{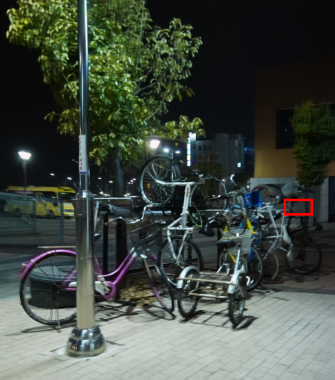}
      \includegraphics[width=\linewidth]{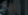}  	
    \end{minipage}
	}\\
\subfloat[FMIMO-UNet+ \cite{mao2023intriguing}]{
    \begin{minipage}{.195\linewidth} 
      \centering
      \includegraphics[width=\linewidth]{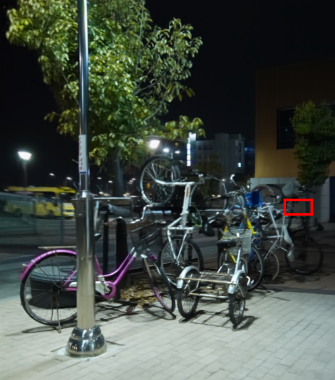}
      \includegraphics[width=\linewidth]{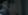}  	
    \end{minipage}
	}
\subfloat[FFTformer \cite{kong2023efficient} ]{
    \begin{minipage}{.195\linewidth} 
      \centering
      \includegraphics[width=\linewidth]{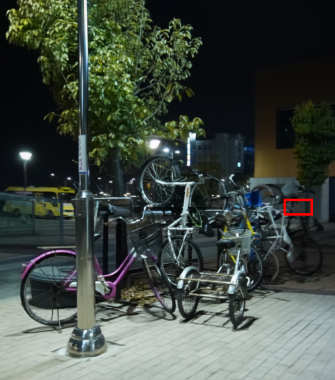}
      \includegraphics[width=\linewidth]{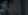}  	
    \end{minipage}
	}
\subfloat[GRL \cite{li2023efficient}]{
    \begin{minipage}{.195\linewidth} 
      \centering
      \includegraphics[width=\linewidth]{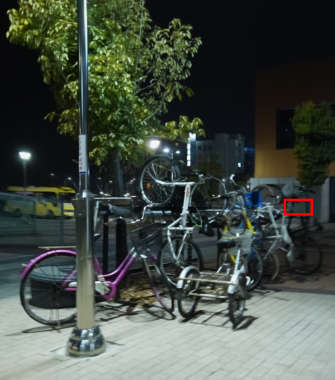}
      \includegraphics[width=\linewidth]{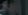}  	
    \end{minipage}
	}
\subfloat[\ours]{
    \begin{minipage}{.195\linewidth} 
      \centering
      \includegraphics[width=\linewidth]{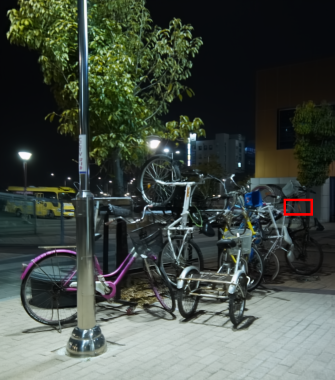}
      \includegraphics[width=\linewidth]{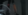}  	
    \end{minipage}
	}
\subfloat[GT]{
    \begin{minipage}{.195\linewidth} 
      \centering
      \includegraphics[width=\linewidth]{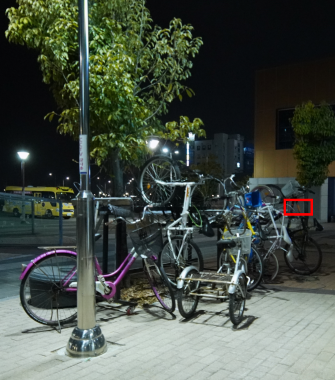}
      \includegraphics[width=\linewidth]{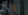}  	
    \end{minipage}
	}\hfill
  \end{minipage}
	\vfill
      \caption{Visual comparisons of single-image motion deblurring on \dataset{RealBlur\_J} \cite{rim2020real}.}
\label{fig:s_realj4}
\end{figure*}
\begin{figure*}[hbt!]
  \centering
  \tiny
  \begin{minipage}[b]{1.\linewidth}
\subfloat[Blurry Input]{
    \begin{minipage}{.195\linewidth} 
      \centering
      \includegraphics[width=\linewidth]{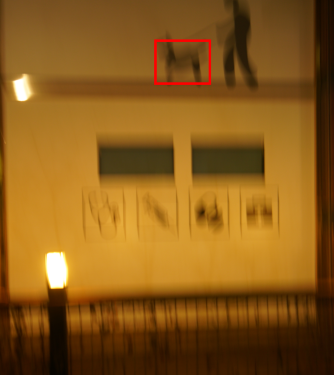}
      \includegraphics[width=\linewidth]{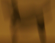}  	
    \end{minipage}
	}
\subfloat[DeblurGAN-v2 \cite{kupyn2019deblurgan}]{
    \begin{minipage}{.195\linewidth} 
      \centering
      \includegraphics[width=\linewidth]{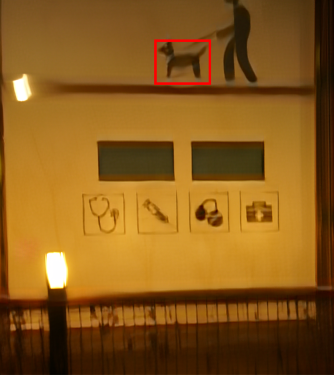}
      \includegraphics[width=\linewidth]{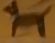}  	
    \end{minipage}
	}
\subfloat[MAXIM-3S \cite{tu2022maxim}]{
    \begin{minipage}{.195\linewidth} 
      \centering
      \includegraphics[width=\linewidth]{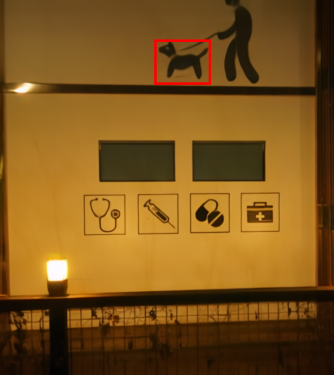}
      \includegraphics[width=\linewidth]{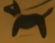}  	
    \end{minipage}
	}
\subfloat[DeepRFT+ \cite{mao2021deep}]{
    \begin{minipage}{.195\linewidth} 
      \centering
      \includegraphics[width=\linewidth]{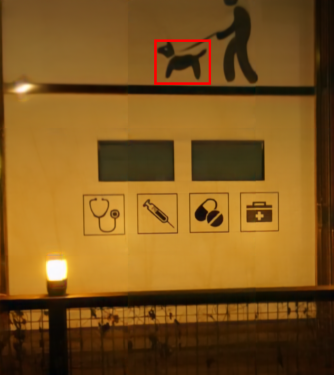}
      \includegraphics[width=\linewidth]{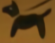}  	
    \end{minipage}
	}
\subfloat[Stripformer \cite{tsai2022stripformer} ]{
    \begin{minipage}{.195\linewidth} 
      \centering
      \includegraphics[width=\linewidth]{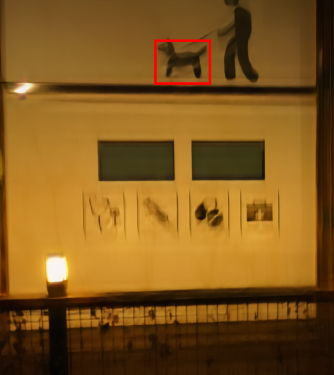}
      \includegraphics[width=\linewidth]{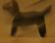}  	
    \end{minipage}
	}\\
\subfloat[FMIMO-UNet+ \cite{mao2023intriguing}]{
    \begin{minipage}{.195\linewidth} 
      \centering
      \includegraphics[width=\linewidth]{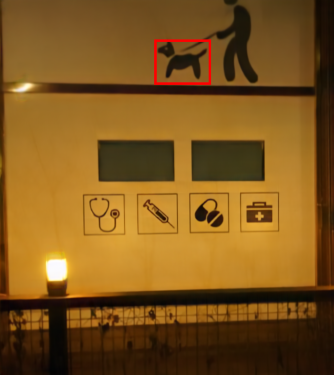}
      \includegraphics[width=\linewidth]{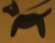}  	
    \end{minipage}
	}
\subfloat[FFTformer \cite{kong2023efficient} ]{
    \begin{minipage}{.195\linewidth} 
      \centering
      \includegraphics[width=\linewidth]{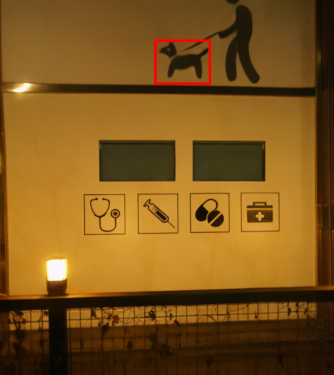}
      \includegraphics[width=\linewidth]{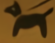}  	
    \end{minipage}
	}
\subfloat[GRL \cite{li2023efficient}]{
    \begin{minipage}{.195\linewidth} 
      \centering
      \includegraphics[width=\linewidth]{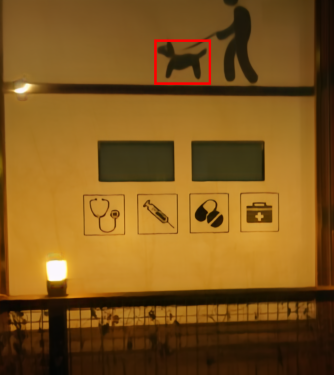}
      \includegraphics[width=\linewidth]{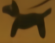}  	
    \end{minipage}
	}
\subfloat[\ours]{
    \begin{minipage}{.195\linewidth} 
      \centering
      \includegraphics[width=\linewidth]{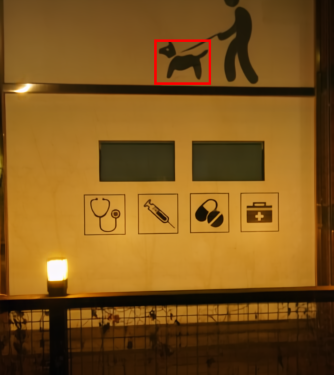}
      \includegraphics[width=\linewidth]{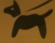}  	
    \end{minipage}
	}
\subfloat[GT]{
    \begin{minipage}{.195\linewidth} 
      \centering
      \includegraphics[width=\linewidth]{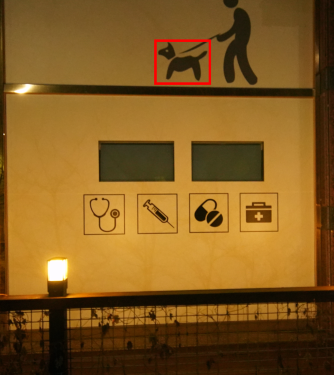}
      \includegraphics[width=\linewidth]{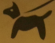}  	
    \end{minipage}
	}\hfill
  \end{minipage}
	\vfill
      \caption{Visual comparisons of single-image motion deblurring on \dataset{RealBlur\_J} \cite{rim2020real}.}
\label{fig:s_realj3}
\end{figure*}
\begin{figure*}[hbt!]
  \centering
  \tiny
  \begin{minipage}[b]{1.\linewidth}
\subfloat[Blurry Input]{
    \begin{minipage}{.195\linewidth} 
      \centering
      \includegraphics[width=\linewidth]{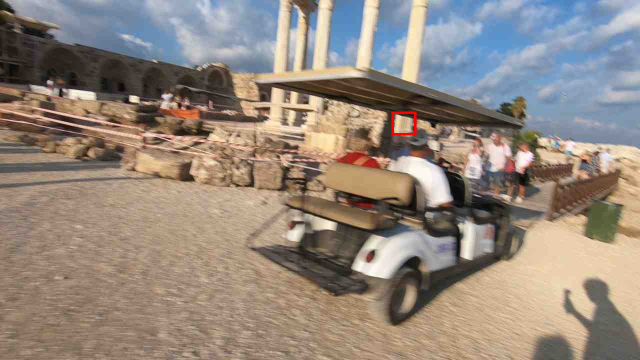}
      \includegraphics[width=\linewidth]{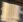}  	
    \end{minipage}
	}
\subfloat[HINet \cite{chen2021hinet}]{
    \begin{minipage}{.195\linewidth} 
      \centering
      \includegraphics[width=\linewidth]{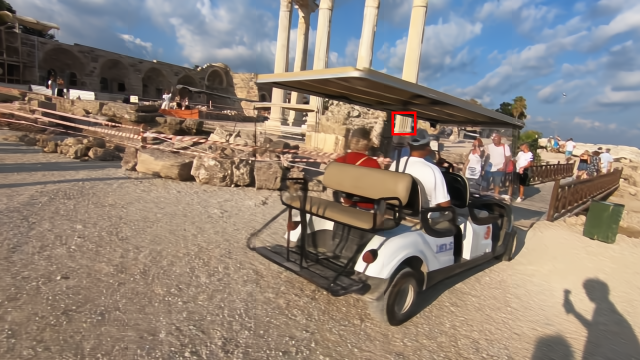}
      \includegraphics[width=\linewidth]{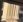}  	
    \end{minipage}
	}
\subfloat[NAFNet \cite{chen2022simple} ]{
    \begin{minipage}{.195\linewidth} 
      \centering
      \includegraphics[width=\linewidth]{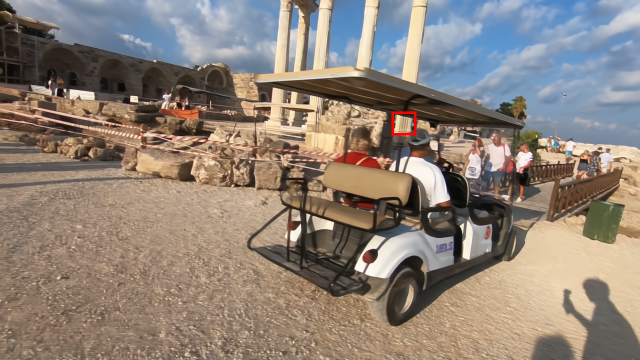}
      \includegraphics[width=\linewidth]{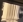}  	
    \end{minipage}
	}
\subfloat[\ours]{
    \begin{minipage}{.195\linewidth}
      \centering
      \includegraphics[width=\linewidth]{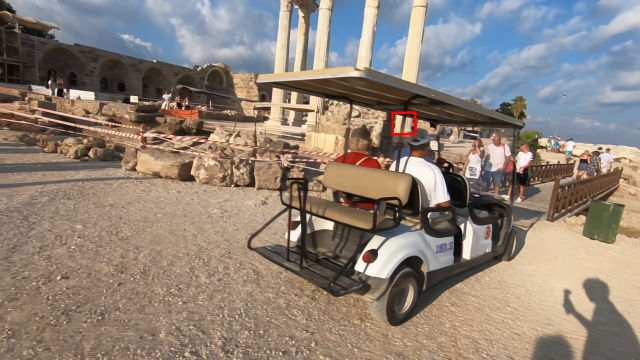}
      \includegraphics[width=\linewidth]{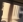}
    \end{minipage}
	}
\subfloat[GT]{
    \begin{minipage}{.195\linewidth} 
      \centering
      \includegraphics[width=\linewidth]{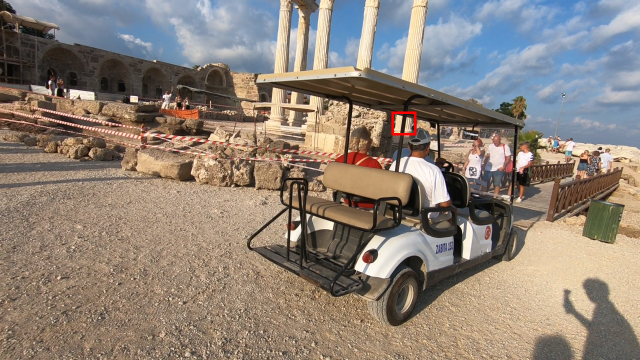}
      \includegraphics[width=\linewidth]{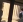}  	
    \end{minipage}
	}\hfill
  \end{minipage}
	\vfill
      \caption{Visual comparisons of motion deblurring with JPEG artifacts on REDS-val-300 \cite{nah2021ntire}.}
\label{fig:reds}
\end{figure*}
\begin{figure*}[hbt!]
  \centering
  \tiny
  \begin{minipage}[b]{1.\linewidth}
\subfloat[Blurry Input]{
    \begin{minipage}{.195\linewidth} 
      \centering
      \includegraphics[width=\linewidth]{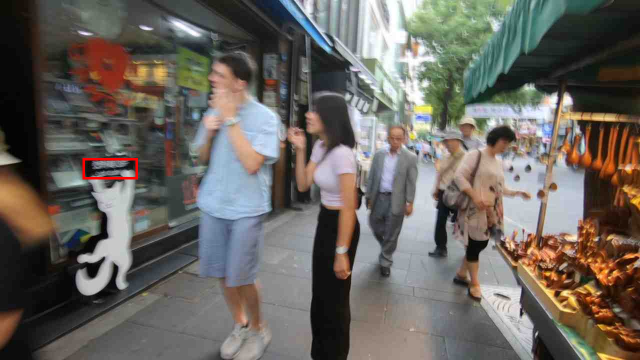}
      \includegraphics[width=\linewidth]{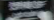}  	
    \end{minipage}
	}
\subfloat[HINet \cite{chen2021hinet}]{
    \begin{minipage}{.195\linewidth} 
      \centering
      \includegraphics[width=\linewidth]{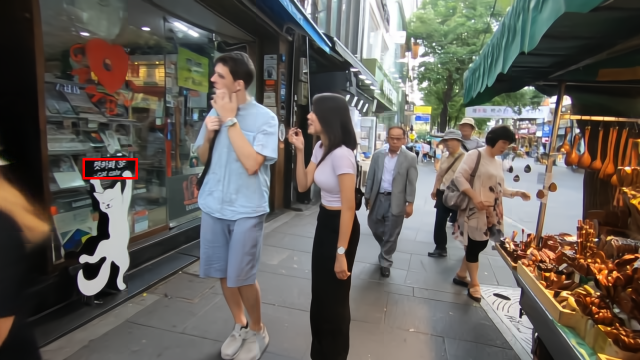}
      \includegraphics[width=\linewidth]{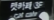}  	
    \end{minipage}
	}
\subfloat[NAFNet \cite{chen2022simple} ]{
    \begin{minipage}{.195\linewidth} 
      \centering
      \includegraphics[width=\linewidth]{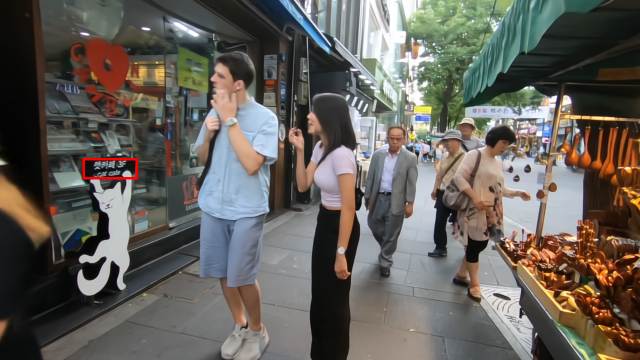}
      \includegraphics[width=\linewidth]{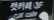}  	
    \end{minipage}
	}
\subfloat[\ours]{
    \begin{minipage}{.195\linewidth}
      \centering
      \includegraphics[width=\linewidth]{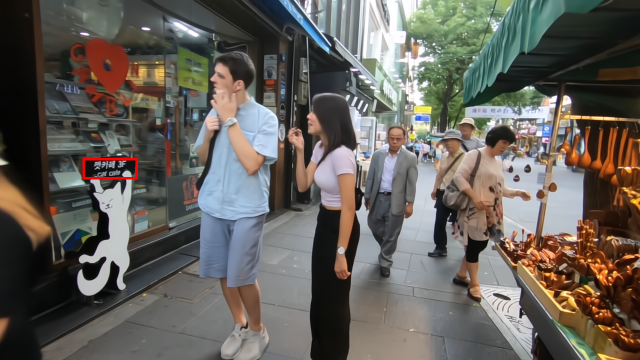}
      \includegraphics[width=\linewidth]{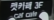}
    \end{minipage}
	}
\subfloat[GT]{
    \begin{minipage}{.195\linewidth} 
      \centering
      \includegraphics[width=\linewidth]{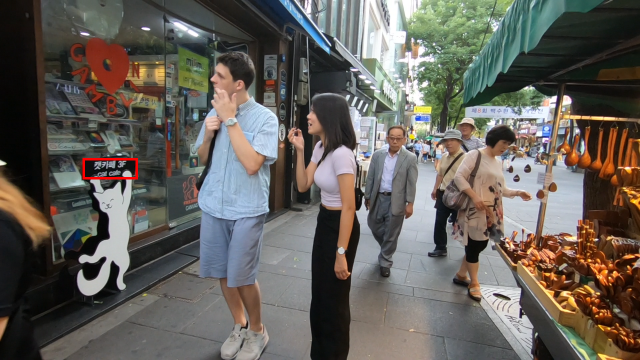}
      \includegraphics[width=\linewidth]{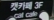}  	
    \end{minipage}
	}\hfill
  \end{minipage}
	\vfill
      \caption{Visual comparisons of motion deblurring with JPEG artifacts on REDS-val-300 \cite{nah2021ntire}.}
\label{fig:reds2}
\end{figure*}
\begin{figure*}
  \centering
  \tiny
  \begin{minipage}[b]{1.\linewidth}
\subfloat[Rainy Input]{
    \begin{minipage}{.195\linewidth} 
      \centering
      \includegraphics[width=\linewidth]{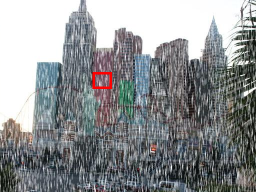}
      \includegraphics[width=\linewidth]{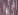}  	
    \end{minipage}
	}
\subfloat[DerainNet \cite{fu2017clearing}]{
    \begin{minipage}{.195\linewidth} 
      \centering
      \includegraphics[width=\linewidth]{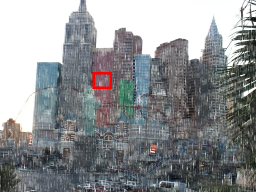}
      \includegraphics[width=\linewidth]{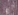}  	
    \end{minipage}
	}
\subfloat[SEMI \cite{wei2019semi} ]{
    \begin{minipage}{.195\linewidth} 
      \centering
      \includegraphics[width=\linewidth]{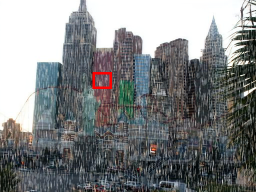}
      \includegraphics[width=\linewidth]{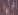}  	
    \end{minipage}
	}
\subfloat[UMRL \cite{yasarla2019uncertainty} ]{
    \begin{minipage}{.195\linewidth} 
      \centering
      \includegraphics[width=\linewidth]{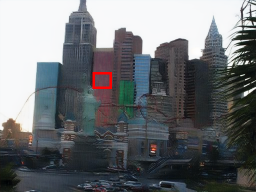}
      \includegraphics[width=\linewidth]{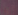}  	
    \end{minipage}
	}
\subfloat[RESCAN \cite{li2018recurrent} ]{
    \begin{minipage}{.195\linewidth} 
      \centering
      \includegraphics[width=\linewidth]{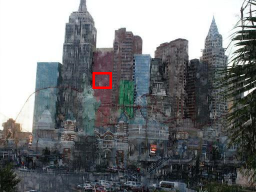}
      \includegraphics[width=\linewidth]{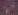}  	
    \end{minipage}
	}\\
\subfloat[PreNet \cite{ren2019progressive} ]{
    \begin{minipage}{.195\linewidth} 
      \centering
      \includegraphics[width=\linewidth]{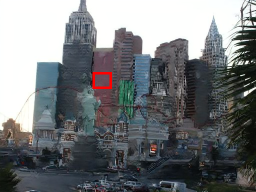}
      \includegraphics[width=\linewidth]{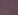}  	
    \end{minipage}
	}
\subfloat[MPRNet \cite{zamir2021multi}]{
    \begin{minipage}{.195\linewidth} 
      \centering
      \includegraphics[width=\linewidth]{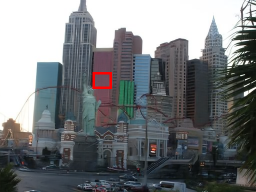}
      \includegraphics[width=\linewidth]{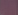}  	
    \end{minipage}
	} 
\subfloat[Restormer \cite{zamir2022restormer}\label{fig:test100res} ]{
    \begin{minipage}{.195\linewidth} 
      \centering
      \includegraphics[width=\linewidth]{figSupp/figDerain/rescan_BOX.png}
      \includegraphics[width=\linewidth]{figSupp/figDerain/rescan_C1.png}  	
    \end{minipage}
	}
\subfloat[\ours]{
    \begin{minipage}{.195\linewidth} 
      \centering
      \includegraphics[width=\linewidth]{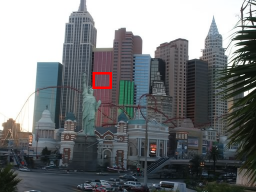}
      \includegraphics[width=\linewidth]{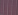}  	
    \end{minipage}
	}
\subfloat[GT]{
    \begin{minipage}{.195\linewidth} 
      \centering
      \includegraphics[width=\linewidth]{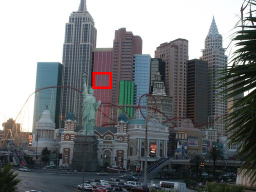}
      \includegraphics[width=\linewidth]{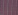}  	
    \end{minipage}
	}\hfill
  \end{minipage}
	\vfill
      \caption{Visual comparisons of deraining on \dataset{Test100} \cite{zhang2019image}.}
\label{fig:test100}
\end{figure*}
\begin{figure*}
  \centering
  \tiny
  \begin{minipage}[b]{1.\linewidth}
\subfloat[Rainy Input]{
    \begin{minipage}{.195\linewidth} 
      \centering
      \includegraphics[width=\linewidth]{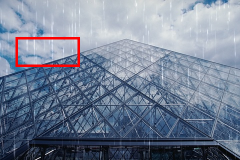}
      \includegraphics[width=\linewidth]{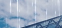}  	
    \end{minipage}
	}
\subfloat[DerainNet \cite{fu2017clearing}]{
    \begin{minipage}{.195\linewidth} 
      \centering
      \includegraphics[width=\linewidth]{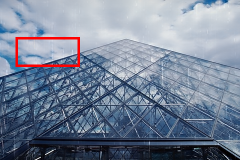}
      \includegraphics[width=\linewidth]{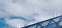}  	
    \end{minipage}
	}
\subfloat[SEMI \cite{wei2019semi} ]{
    \begin{minipage}{.195\linewidth} 
      \centering
      \includegraphics[width=\linewidth]{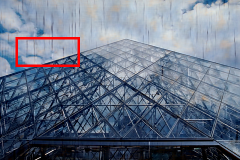}
      \includegraphics[width=\linewidth]{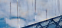}  	
    \end{minipage}
	}
\subfloat[UMRL \cite{yasarla2019uncertainty} ]{
    \begin{minipage}{.195\linewidth} 
      \centering
      \includegraphics[width=\linewidth]{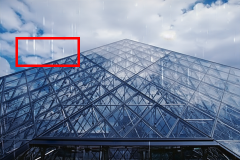}
      \includegraphics[width=\linewidth]{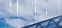}  	
    \end{minipage}
	}
\subfloat[RESCAN \cite{li2018recurrent} ]{
    \begin{minipage}{.195\linewidth} 
      \centering
      \includegraphics[width=\linewidth]{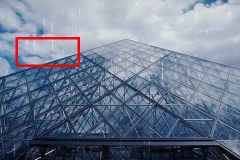}
      \includegraphics[width=\linewidth]{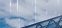}  	
    \end{minipage}
	}\\
\subfloat[PreNet \cite{ren2019progressive} ]{
    \begin{minipage}{.195\linewidth} 
      \centering
      \includegraphics[width=\linewidth]{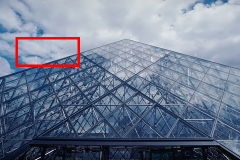}
      \includegraphics[width=\linewidth]{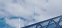}  	
    \end{minipage}
	}
\subfloat[MPRNet \cite{zamir2021multi}]{
    \begin{minipage}{.195\linewidth} 
      \centering
      \includegraphics[width=\linewidth]{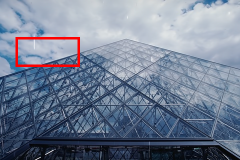}
      \includegraphics[width=\linewidth]{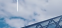}  	
    \end{minipage}
	} 
\subfloat[Restormer \cite{zamir2022restormer}\label{fig:rain100Lres} ]{
    \begin{minipage}{.195\linewidth} 
      \centering
      \includegraphics[width=\linewidth]{figSupp/figDerain2/rescan_BOX.png}
      \includegraphics[width=\linewidth]{figSupp/figDerain2/rescan_C1.png}  	
    \end{minipage}
	}
\subfloat[\ours]{
    \begin{minipage}{.195\linewidth} 
      \centering
      \includegraphics[width=\linewidth]{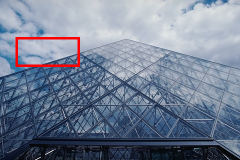}
      \includegraphics[width=\linewidth]{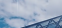}  	
    \end{minipage}
	}
\subfloat[GT]{
    \begin{minipage}{.195\linewidth} 
      \centering
      \includegraphics[width=\linewidth]{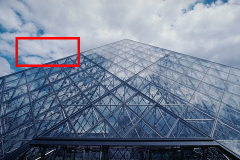}
      \includegraphics[width=\linewidth]{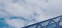}  	
    \end{minipage}
	}\hfill
  \end{minipage}
	\vfill
      \caption{Visual comparisons of deraining on \dataset{Rain100L} \cite{yang2017deep}.}
\label{fig:rain100l}
\end{figure*}

\end{document}